\documentclass[10pt,journal,compsoc]{IEEEtran}
\ifCLASSOPTIONcompsoc
  \usepackage[nocompress]{cite}
\else
  \usepackage{cite}
\fi
\ifCLASSINFOpdf
\else
\fi
\usepackage{amsmath}
\usepackage{amssymb}
\usepackage{graphicx}

\usepackage{epsfig}

\usepackage{url}
\usepackage{booktabs}
\usepackage{makecell}
\usepackage{multirow}
\usepackage{caption}
\usepackage{gensymb}
\usepackage[linesnumbered,lined,boxed,commentsnumbered,ruled]{algorithm2e}
\usepackage[colorlinks,linkcolor=black,anchorcolor=black,citecolor=black,urlcolor=black]{hyperref}
\usepackage{rotating}
\usepackage[caption=false]{subfig}
\renewcommand{\paragraph}[1]{\noindent\textbf{#1}~~}

\usepackage{subfloat}
\usepackage{tabularx}

\ifCLASSOPTIONcompsoc
  \usepackage[nocompress]{cite}
\else
  \usepackage{cite}
\fi

\usepackage{algorithmic}

\usepackage{array}

\usepackage{url}

\begin{document}
\title{TransVOD: End-to-End Video Object Detection \\ with Spatial-Temporal Transformers}

\author{Qianyu~Zhou\textsuperscript{$\dagger$},
        Xiangtai~Li\textsuperscript{$\dagger$},
        Lu~He\textsuperscript{$\dagger$},
        Yibo Yang, 
        Guangliang~Cheng, \\
        Yunhai Tong, 
        Lizhuang~Ma,
        Dacheng~Tao, ~\textit{Fellow}, IEEE
        % <-this % stops a space
% \IEEEcompsocitemizethanks{\IEEEcompsocthanksitem M. Xu, and H. Wang and B. Ni are with the school of Shanghai Jiao Tong University, Shanghai 200240, China.\protect\\
\IEEEcompsocitemizethanks{
%\IEEEcompsocthanksitem Manuscript received xx xx, 2021. revised xx xx 2021.

\IEEEcompsocthanksitem Qianyu Zhou, Lu He and Lizhuang Ma  are with the Department of Computer Science and Engineering, Shanghai Jiao Tong University, Shanghai 200240, China (E-mail: \{zhouqianyu, 147258369\}@sjtu.edu.cn, ma-lz@cs.sjtu.edu.cn).  \protect
\IEEEcompsocthanksitem Xiangtai Li and Yunhai Tong are with the School of Artificial Intelligence, Peking University, Beijing 100871, China
(E-mail: \{lxtpku, yhtong\}@pku.edu.cn). 
\IEEEcompsocthanksitem Yibo Yang and Dacheng Tao are with JD Explore Academy, Beijing 100176, China (E-mail: ibo@pku.edu.cn, dacheng.tao@sydney.edu.au)
\IEEEcompsocthanksitem Guangliang Cheng is with the SenseTime Research, Beijing 100080, China (E-mail: guangliangcheng2014@gmail.com).
\IEEEcompsocthanksitem \textsuperscript{$\dagger$} indicates the first three authors have equal contributions.

Manuscript received 10 March 2022; revised 14 September 2022; accepted 9 November 2021. Date of publication XX 2023; date of current version 18 November 2022.

This work is supported by National Key Research and Development Program of China (2019YFC1521104), National Natural Science Foundation of China (72192821, 61972157), Shanghai Municipal Science and Technology Major Project  (2021SHZDZX0102), Shanghai Science and Technology Commission (21511101200, 22YF1420300), and Art major project of National Social Science Fund (I8ZD22). 

(Corresponding authors: Lizhuang Ma and Xiangtai Li.)

Recommended for acceptance by XXX.

Digital Object Identifier no. 10.1109/TPAMI.2022.3223955.
%\IEEEcompsocthanksitem 
%\IEEEcompsocthanksitem 
}% <-this % stops an unwanted space
% \thanks{Manuscript received April 19, 2005; revised August 26, 2015.}
}

\markboth{IEEE TRANSACTIONS ON PATTERN ANALYSIS AND MACHINE INTELLIGENCE,~Vol.~X, No.~X, X}%
{Shell \MakeLowercase{\textit{et al.}}: Bare Demo of IEEEtran.cls for IEEE Journals}

\IEEEtitleabstractindextext{
\begin{abstract}

Detection Transformer (DETR) and Deformable DETR have been proposed to eliminate the need for many hand-designed components in object detection while demonstrating good performance as previous complex hand-crafted detectors. However, their performance on Video Object Detection (VOD) has \emph{not} been well explored. In this paper, we present \textbf{TransVOD}, the \emph{first end-to-end} video object detection system based on simple yet effective spatial-temporal Transformer architectures. The first goal of this paper is to streamline the pipeline of current VOD, effectively removing the need for many hand-crafted components for feature aggregation, \emph{e.g.,} optical flow model, relation networks. Besides, benefited from the object query design in DETR, our method does not need post-processing methods such as Seq-NMS. In particular, we present a temporal Transformer to aggregate both the spatial object queries and the feature memories of each frame. Our temporal transformer consists of two components: Temporal Query Encoder (TQE) to fuse object queries, and Temporal Deformable Transformer Decoder (TDTD) to obtain current frame detection results. These designs boost the strong baseline deformable DETR by a significant margin (3 \%-4 \% mAP) on the ImageNet VID dataset. TransVOD yields comparable performances on the benchmark of ImageNet VID. Then, we present two improved versions of TransVOD including TransVOD++ and TransVOD Lite. The former fuses object-level information into object query via dynamic convolution while the latter models the entire video clips as the output to speed up the inference time. We give detailed analysis of all three models in the experiment part. In particular, our proposed TransVOD++ sets a new state-of-the-art record in terms of accuracy on ImageNet VID with 90.0 \% mAP. Our proposed TransVOD Lite also achieves the best speed and accuracy trade-off with 83.7 \% mAP while running at around 30 FPS on a single V100 GPU device. Code and models are available at \href{https://github.com/SJTU-LuHe/TransVOD}{\textit{https://github.com/SJTU-LuHe/TransVOD}}.
\end{abstract}

}

\maketitle
% \IEEEdisplaynontitleabstractindextext

\IEEEpeerreviewmaketitle

\section{Introduction}

\begin{figure}[t]
    \centering
    \includegraphics[width=1.0\linewidth]{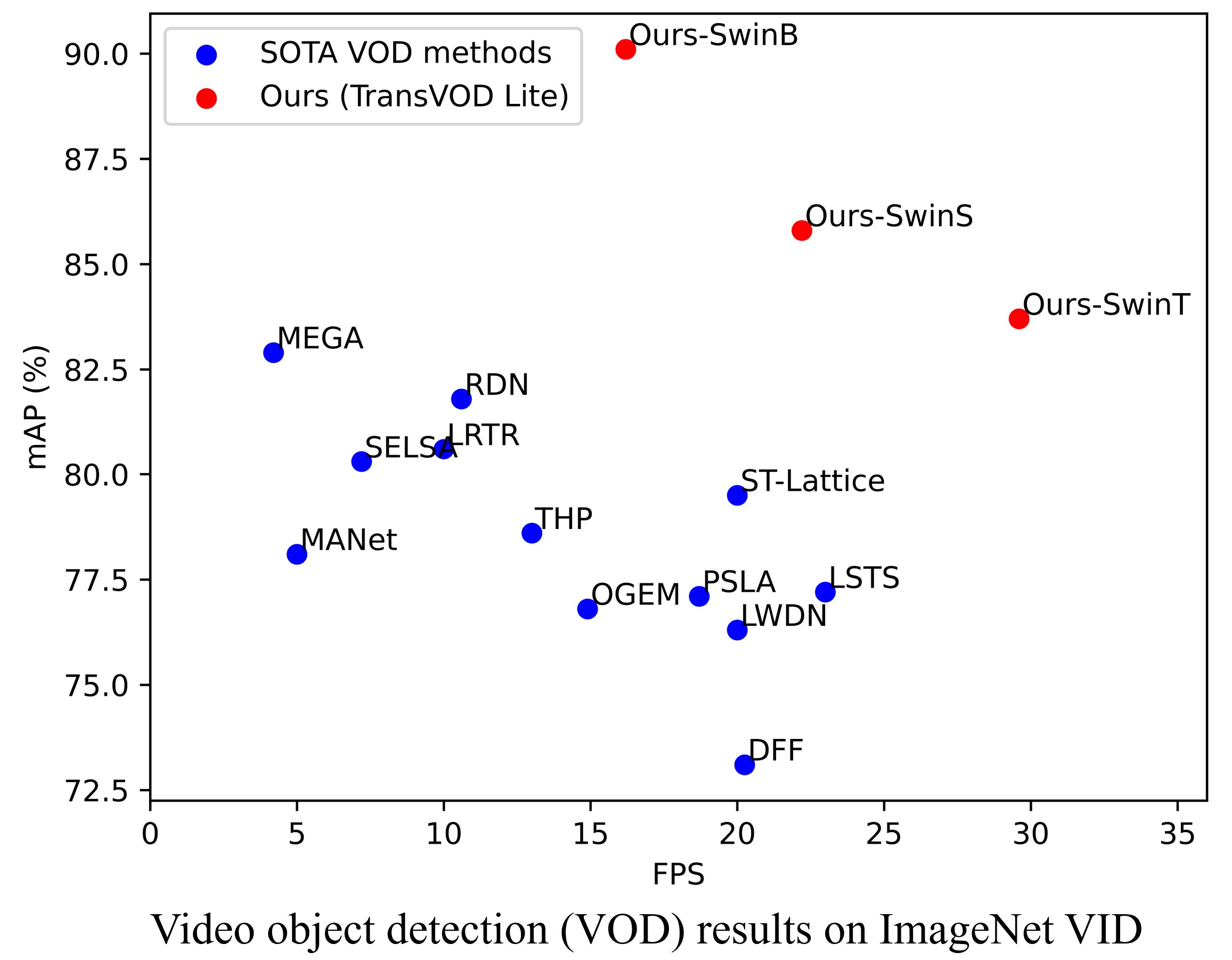} 
    \caption{\small Speed and Accuracy trade-off of video object detection (VOD) results in ImageNet VID dataset. The blue points plot the state-of-the-art (SOTA) VOD methods, and the red ones are our proposed method TransVOD Lite, achieving the \textbf{best} trade-off between the speed and accuracy with different backbones. SwinB, SwinS and SwinT mean Swin Base, Small and Tiny. }
     \label{fig:teaser_speed_acc}
     \vspace{-5mm}
\end{figure}

\IEEEPARstart{V}{ideo} Object Detection (VOD) extends image object detection to video scenarios, which aims to detect every object given video clips. 
It enables various applications in the real world, \emph{e.g.,} autonomous driving. However, still-image detectors~\cite{ren2016faster,dai16rfcn,FocalLoss,tian2019fcos} cannot be directly applied to much challenging video data, due to the appearance deterioration and changes of video frames, \emph{e.g.,} motion blur, part occlusion, camera refocous and rare poses.

Previous VOD methods mainly leverage the temporal information in two different manners. The first one relies on post-processing of temporal information~\cite{han2016seq,kang2017t,belhassen2019improving,sabater2020robust} to make the detection results more coherent and stable. These methods usually apply a still-image detector to obtain detection results, then associate the results. Another line of approaches~\cite{yao2020video,jiang2020learning,han2020mining,han2020exploiting,lin2020dual,he2020temporal,chen2018optimizing,chen2020memory,sun2021mamba} exploits the feature aggregation of temporal information. Specifically, they improve features of the current frame by aggregating that of adjacent frames or entire clips to boost the detection performance via specific operator design. In this way, the problems such as motion blur, part occlusion, and fast appearance change can be well solved. In particular, most methods~\cite{chen2018optimizing,chen2020memory,sun2021mamba,guo2019progressive} use two-stage detector Faster-RCNN~\cite{ren2016faster} or R-FCN~\cite{dai16rfcn} as the still-image baseline.

\begin{figure*}[t]
    \centering
    \includegraphics[width=1.0\textwidth]{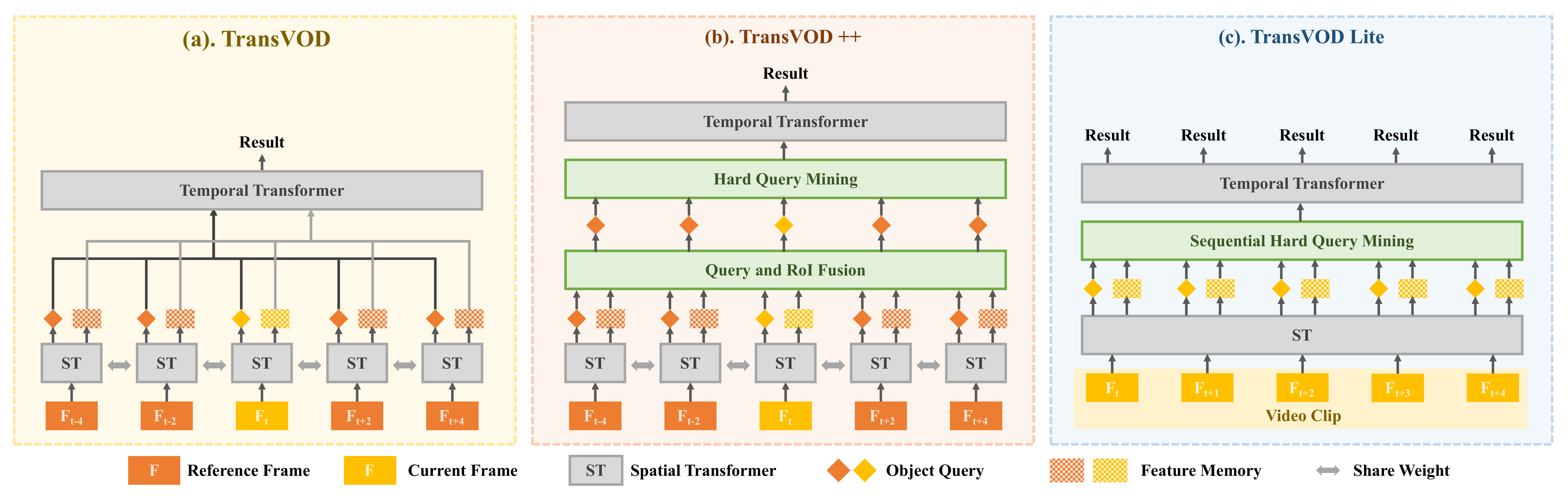} 
    \caption{\small Illustration of our proposed TransVOD series. (a) Original TransVOD: our network is based on spatial Transformers which outputs spatial object query and feature memory of each frame. We propose a temporal Transformer to link both the spatial object queries and feature memories in a temporal dimension to obtain the results of the current frame. The final detection results are obtained via a shared feed-forward network (FFN).
    (b) Based on TransVOD, our TransVOD++ add two improvements including Hard Query Mining (HQM) and Query and RoI Fusion module (QRF). (c) Inherited from TransVOD, our TransVOD Lite models the VOD task as a sequence-to-sequence prediction problem, and \textbf{directly} outputs all the detection results of the entire sequence in the window via Sequential Hard Query Mining (SeqHQM).}
    \label{fig:teaser_arch}
    \vspace{-3mm}
\end{figure*}

Despite the gratifying success of these approaches, most of the two-stage pipelines for video object detection are over sophisticated, requiring many hand-crafted components, \emph{e.g.,} optical flow model~\cite{zhu17dff,zhu17fgfa,wang18manet,zhu18hp,jin2022feature}, recurrent neural network~\cite{deng2019ogemn,chen2020memory,guo2019progressive}, deformable convolution fusion~\cite{bertasius18stsn,jiang2019video,he2020temporal}, relation networks~\cite{deng19rdn,chen2020memory,shvets19lltr}. In addition, most of them need complicated post-processing methods by linking the same object across the video to form tubelets and aggregating classification scores in the tubelets to achieve the state-of-the-art performance~\cite{han2016seq,kang2017t,belhassen2019improving,sabater2020robust}. 
Meanwhile, there are also several studies~\cite{liu2019looking,liu2018mobile,Chen2018OptimizingVOD,jiang2019video,jiang2020learning,yao2020video} focusing on real-time video object detection. However, these works still need sophisticated designs. Thus, it is \textit{in desperate need to build \textbf{a simple yet effective} VOD framework in a fully end-to-end manner}. 

Transformers~\cite{wang2020end,detr,zhu2020deformable,dosovitskiy2020image,sun2020transtrack} have shown promising potential in computer vision. Especially, DETR~\cite{detr,zhu2020deformable} simplifies the detection pipeline by modeling the object queries and achieving comparative performance with highly optimized CNN-based detectors. However, such static detectors cannot handle motion blur, part occlusion, video defocus, or rare poses well due to the lack of temporal information, which will be shown in the experiment part. Thus, how to model the temporal information in a long-range video clip is a very critical problem. 

In this paper, our goal is to extend the DETR-like object detector into the video object detection domain. Our insights are four aspects. \textit{Firstly}, we observe that the video clip contains rich inherent temporal information, \emph{e.g.,} rich visual cues of motion patterns. Thus, it is natural to view video object detection as a sequence-to-sequence task with the advantages of Transformers~\cite{Vaswani17attention}. 
The whole video clip is like a sentence, and each frame contributes similarly to each word in natural language processing. Transformers can not only be used in inner each frame to model the interaction of each object, but also be used to link objects along the temporal dimension. Secondly, object query is one key component design in DETR~\cite{detr} which encodes instance-aware information. The learning process of DETR can be seen as the grouping process: grouping each object into an object query. Thus, these query embeddings can represent the instances of each frame, and it is natural to link these sparse query embeddings via another temporal transformer. Thirdly, the output memory from the DETR transformer encoder contains rich spatial information which can also be modeled jointly with query embeddings along the temporal dimension. Fourthly, adopting clip-level inputs of Transformers can speed up the object detection process in a video, which is needed in many real-world applications. 

Motivated by the above facts, we propose TransVOD, a novel end-to-end video object detection framework based on a spatial-temporal Transformer architecture. Our TransVOD views video object detection as an end-to-end sequence decoding/prediction problem. For the current frame, as shown in Fig.~(\ref{fig:teaser_arch})(a), it takes multiple frames as inputs and directly outputs the current frame detection results via a Transformer-like architecture. In particular, we design a novel temporal Transformer to link each object query and outputs of memory encodings simultaneously. Our proposed temporal Transformer mainly contains three components: Temporal Deformable Transformer Encoder (TDTE) to encode the multiple frame spatial details, Temporal Query Encoder (TQE) to fuse object queries in one video clip, and Temporal Deformable Transformer Decoder (TDTD) to obtain the final detection results of the current frame. TDTE efficiently aggregates the spatial information via temporal deformable attention and avoids the background noises. TQE first adopts a coarse-to-fine strategy to select relevant object queries in one clip and fuse such selected queries via several self-attention layers~\cite{Vaswani17attention}. TDTD is another decoder that takes the outputs of TDTE and TQE as inputs, and directly outputs the final detection results. These modules are shared for each frame and can be trained in an fully end-to-end manner. We carry out extensive experiments on ImageNet VID dataset~\cite{russakovsky2015imagenet}. Compared with the single-frame baseline~\cite{zhu2020deformable}, our TransVOD achieves significant improvements (2\%$\sim$4\% mAP). 
% This preliminary work is published in ACM MM 2021~\cite{he2021end}.

Based on the TransVOD framework, which is published in ACM MM 2021~\cite{he2021end}, we present two improved versions including TransVOD++ and TransVOD Lite. For TransVOD++, regarding that 
there exists large redundancy in both the number of object queries and the targets, we present a hard query mining (HQM) strategy to sample the hardest queries during the training inspired from the hard pixels mining in image object detection and segmentation~\cite{FocalLoss,ohem,SegOHEM}, as shown in Fig.~\ref{fig:teaser_arch}(b). Moreover, we present a novel query and RoI fusion (QRF) module via dynamic convolutions. In this way, the object-level appearance information is injected into each query and TDTE can be avoided since the spatial fusion can be replaced with QRF. Compared with previous TransVOD, we find both improvements lead to better results with faster speed. Moreover, when deploying the vision Transformer backbone~\cite{liu2021swin}, we present a simply-aligned fusion to fuse multi-scale features for TDTD. After adopting Swin base as the backbone, our TransVOD++ achieves 90\% mAP on the ImageNet VID dataset and suppress previous works by a significant margin (5\%$\sim$6 \%) with a simpler pipeline. Our method is \textbf{the first to achieve 90\% mAP on ImageNet VID dataset}.

Inherited from TranVOD, we present TransVOD Lite, aiming at real-time VOD and modeling the VOD task as a sequence-to-sequence prediction problem which is adopted in machine translation~\cite{Vaswani17attention}. The pipeline is shown in Fig.~\ref{fig:teaser_arch}(c). In particular, given a window size $T$ ($T$ can be chosen in {8, 16}), we take multiple frames as inputs and obtain multiple frame results simultaneously. Then, one video clip results can be obtained in a temporal window manner. In this way, we can fully use the memory of GPU to speed up inference time. Our TransVOD Lite can boost the single image baseline by 2\%$\sim$3\% mAP but with a faster speed (4x-6x). After adopting the Swin Transformer, as shown in Fig.~\ref{fig:teaser_speed_acc}, our methods achieve the best speed and accuracy trade-off. Our methods lead to a significant margin (3\%$\sim$4\%mAP, 5$\sim$15 FPS) compared with previous VOD methods in both speed and accuracy. Our best model can achieve 83.7\% mAP while running at around 30 FPS. In summary, following the TransVOD framework, we present TransVOD++ and TransVOD Lite. Both models set new state-of-the-art results on the challenging ImageNet VID dataset in two different settings: accuracy for non-real-time models and best speed-accuracy trade-off on real-time models. These results indicate our method can be new solid baseline for VOD.

\section{Related work}
\label{sec:related}
\noindent
\textbf{Video Object Detection.} VOD task requires detecting objects in each frame and linking the same objects across frames. State-of-the-art methods typically develop sophisticated pipelines to tackle it. In general, VOD task can be divided into two directions: \textit{improving detection accuracy via temporal fusing} and \textit{performing real-time video object detection while keeping the accuracy}. 

For the first aspect, most previous works~\cite{yao2020video,jiang2020learning,han2020mining,han2020exploiting,lin2020dual,he2020temporal,chen2018optimizing,chen2020memory,sun2021mamba,guo2019progressive,zhu17fgfa,zhu18hp,wang18manet} to amend this problem is feature aggregation that enhances per-frame features by aggregating the features of nearby frames. Earlier works adopt flow-based warping to achieve feature aggregation. 
Specifically, FGFA~\cite{zhu17fgfa} and THP~\cite{zhu18hp} both utilize the optic flow from FlowNet~\cite{dosovitskiy2015flownet} to model the motion relation via different temporal feature aggregation strategies. To calibrate the pixel-level features with inaccurate flow estimation, MANet~\cite{wang18manet} dynamically combines pixel-level and instance-level calibration according to the motion. Nevertheless, these flow-warping-based methods have several disadvantages: 1) Training a model for flow extraction requires large amounts of flow data, which may be difficult and costly to obtain. 2) integrating a flow network and a detection network into a single model may be challenging due to multitask learning. Another line of attention-based approaches~\cite{wu19selsa,he2020temporal,bertasius18stsn,jiang2019video,deng2019ogemn,chen2020memory} utilize self-attention~\cite{vaswani2017attention} and non-local~\cite{wang2018non} to capture long-range dependencies of temporal contexts. SELSA~\cite{wu19selsa} treats video as a bag of unordered frames and proposes to aggregate features in the full-sequence level. STSN~\cite{bertasius18stsn} and TCENet~\cite{he2020temporal} propose to utilize deformable convolution to aggregate the temporal contexts within a complicated framework with so many heuristic designs. RDN~\cite{deng19rdn} introduces a new design to capture the interactions across the objects in spatial-temporal context. LWDN~\cite{jiang2019video} adopts a memory mechanism to propagate and update the memory feature from key frames to key frames. OGEMN~\cite{deng2019ogemn} present to use object-guided external memory to store the pixel and instance-level features for further global aggregation. MEGA~\cite{chen2020memory} considers aggregating both the global information and local information from the video and presents a long-range memory. Despite the great success of these approaches, most of the pipelines for VOD are too sophisticated, requiring many hand-crafted components, \emph{e.g.,} extra optic flow model, memory mechanism, or recurrent neural network. In addition, most of them need complicated post-processing methods such as Seq-NMS~\cite{han2016seq}, Tubelet rescoring~\cite{kang2017t}, Seq-Bbox Matching~\cite{belhassen2019improving}  or REPP~\cite{sabater2020robust} by linking the same object across the video to form tubelets and aggregating classification scores in the tubelets to achieve the state-of-the-art. Instead, our previous work TransVOD builds a \textit{simple and end-to-end trainable} VOD framework without these designs. Beyond that, our improved version TransVOD++ incorporates more appearance information into the object query design and simplifies the whole pipeline by removing the temporal encoder (TDTE) of original TransVOD. It achieves better results than TransVOD and the state-of-the-art performances on the ImageNet VID dataset.

For the second aspect, starting from DFF~\cite{zhu17dff}, several works~\cite{liu2019looking,liu2018mobile,Chen2018OptimizingVOD,jiang2019video,jiang2020learning,yao2020video,chin2019adascale,li2021improving} focus on real-time video object detection while keeping accuracy unchanged or even improved. In general, most of these works also perform specific architecture designs with many hand-crafted components and human prior such as object-level tracker in~\cite{yao2020video}, patchwork cell with attention in~\cite{Chai2019PatchworkAP} and Convolutional LSTMs in~\cite{liu2018mobile}. Our proposed TransVOD Lite models the entire VOD pipeline as a sequence to sequence problem, as Transformer did in machine translation~\cite{Vaswani17attention}. It achieves significant improvements over the strong image baseline along with a faster speed.
\begin{figure*}[t!]
    \centering
    \includegraphics[width=1.0\textwidth]{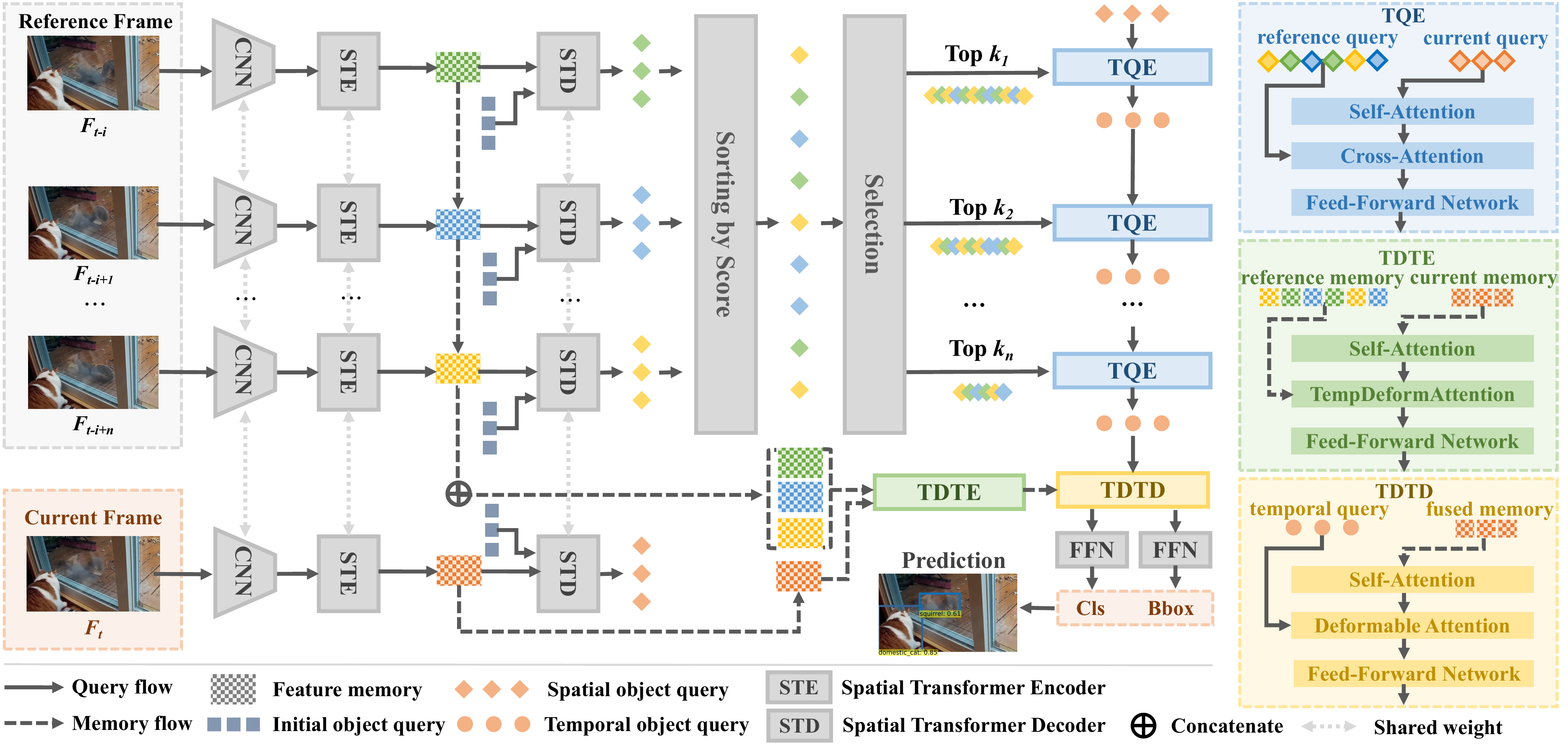} 
    \caption{
    \small \textbf{The whole pipeline of TransVOD.} A shared CNN backbone extracts features of multiple frames. Next, a series of shared Spatial Transformer Encoders (STE) produce the feature memories and these memories are linked and fed into Temporal Deformable Transformer Encoder (TDTE). Meanwhile, the Spatial Transformer Decoder (STD) decodes the spatial object queries. Naturally, we use a Temporal Query Encoder (TQE) to model the relations of different queries and aggregate these queries, thus we can enhance the object query of the current frame. Both the temporal object query and the temporal feature memories are fed into the Temporal Deformable Transformer Decoder (TDTD) to learn the temporal contexts across different frames. Our TransVOD framework can be trained in a fully end-to-end manner.}
    \label{fig:framework_transvod}
    \vspace{-5mm}
\end{figure*}

\noindent
\textbf{Vision Transformers.} Recently, vision Transformers~\cite{detr,zhu2020deformable,liu2021swin, dosovitskiy2020image,sun2020transtrack,meinhardt2021trackformer} make a great progress. It can be mainly divided into two directions: replacing CNN backbone with Transformer-Like architecture~\cite{dosovitskiy2020image,liu2021swin,deit_vit,zhang2022eatformer} and using object query to represent instance for scene understanding~\cite{detr,zhu2020deformable,video_knet,panopticpartformer,fashionformer}. Our work is related to the second part. DETR~\cite{detr} builds a fully end-to-end object detection system based on Transformers, which largely simplifies the traditional detection pipeline. It also achieves on par performances compared with highly-optimized CNN-based detectors~\cite{ren2016faster}. However, it suffers from slow convergence and limited feature spatial resolution, Deformable DETR~\cite{zhu2020deformable} improves DETR by designing a deformable attention module, which attends to a small set of sampling locations as a pre-filter for prominent key elements out of all the feature map pixels. Our work is inspired by DETR~\cite{detr} and Deformable DETR~\cite{zhu2020deformable}. The above works show the effectiveness of Transformers in image object detection tasks. There are several con-current works that applied Transformer into video understanding, \emph{e.g.,} video instance segmentation (VIS)~\cite{vis_dataset}, multi-object tracking (MOT). TransTrack~\cite{sun2020transtrack} introduces a query-key mechanism into the multi-object tracking model, while Trackformer~\cite{meinhardt2021trackformer} directly adds track query for MOT. However, both only leverage limited temporal information, \emph{i.e.,} just the previous frame. We suppose that this way can not fully use enough temporal contexts from a video clip. VisTR~\cite{wang2020end} views the VIS task as a direct end-to-end parallel sequence prediction problem. The targets of a clip are disrupted in such an instance sequence, and directly performing target assignment is not optimal. Instead, we aim to link the outputs of the spatial Transformer, \emph{i.e.,} object query, through a temporal Transformer, which acts in a completely different way from VisTR~\cite{wang2020end}. To our knowledge, there are no prior applications of Transformers to video object detection (VOD) tasks so far. It is intuitive to see that the Transformers’ advantage of modeling long-range dependencies in learning temporal contexts across multiple frames for VOD task. Our previous work, TransVOD~\cite{he2021end}, leverages both the spatial Transformer and the temporal Transformer, and then provide an affirmative answer to that. In this paper, based on the TransVOD framework, we provide two extra solutions including TransVOD++ and TransVOD Lite. The former aims to improve the performance of TransVOD while keeping inference efficiency, while the latter carry out real-time VOD detection with much faster inference speed.

\section{Method}
\label{sec:method}

\noindent
\textbf{Overview.} We first review the previous works, including both DETR~\cite{detr} and Deformable DETR~\cite{zhu2020deformable} in Sec.~\ref{sec:DETR}. Then, we give detailed descriptions of our proposed TransVOD framework in Sec.~\ref{sec:TransVOD}. It contains three key components: Temporal Deformable Transformer Encoder (TDTE), Temporal Query Encoder (TQE), and Temporal Deformable Transformer Decoder (TDTD). Then, we present two advanced versions of our TransVOD including TransVOD++ (Sec.~\ref{sec:TransVOD++} ) and TransVOD Lite (Sec.~\ref{sec:TransVOD_lite}). Finally, we describe the loss functions and details of inference in Sec.~\ref{sec:loss_inference}. 

% \vspace{-2mm}
\subsection{Revisiting DETR and Deformable DETR}
\label{sec:DETR}
DETR~\cite{detr} treats object detection as a set prediction problem. A CNN backbone~\cite{he16res} extracts visual feature maps $f  \in \mathbb{R}^{C \times H \times W}$ from an image and $H,W$ are the height and width of the visual feature map, respectively. The visual features augmented with position embedding $f_{pe}$ would be fed into the encoder of the Transformer. Self-attention would be applied to $f_{pe}$ to generate the key, query, and value features $K, Q, V$ to exchange information between features at all spatial positions.
Let $\Omega_q$ and $\Omega_k$ indicate the set of query and key elements, respectively. Then, $q\in\Omega_q$ denotes the query element and $k\in\Omega_k$ denotes the key element, respectively, which indexes the query feature $z_q \in R^{C}$, and key feature $x_k \in R^{C}$, where $C$ denotes the dimension of the feature.  Then, the multi-head attention feature is as follows:
\begin{align}
    \text{MultiHeadAttn}(z_q, x) = \sum_{m=1}^{M} W_m \big[\sum_{k\in\Omega_k} A_{mqk} \cdot W'_m x_k \big],
    \label{eq:co-attention} 
\end{align}
where $m$ indexes the attention head, $W'_m \in R^{C_v \times C}$ and $W_m \in R^{C \times C_v}$ are learnable weights ($C_v = C/M$ by default). The attention weights $A_{mqk}$ are normalized as:
\begin{align}
   A_{mqk} \propto \exp\{\frac{z_q^T U_m^T~ V_m x_k}{\sqrt{C_v}}\}, \sum_{k\in\Omega_k} A_{mqk} = 1,
\end{align}
where $U_m, V_m \in R^{C_v \times C}$ are learnable weights. The features $z_q$ and $x_k$ are the concatenation/summation of element contents and positional embeddings in practice. 
The decoder's output features of each object query are then further transformed by a Feed-Forward Network (FFN) to output class score and box location for each object. Given box and class prediction, the Hungarian algorithm is applied between predictions and ground-truth box annotations to identify the learning targets of each object query for one-to-one matching. Deformable DETR~\cite{zhu2020deformable} replaces the multi-head self-attention layer with a deformable attention layer to efficiently sample local pixels rather than all pixels. Moreover, to handle missing small objects, they also propose a cross-attention module that incorporates multi-scale feature representations. Due to the fast convergence and computation efficiency, we adopt Deformable DETR~\cite{zhu2020deformable} as our still image Transformer detector.

% \vspace{-2mm}
\subsection{TransVOD Framework} 
\label{sec:TransVOD}
The overall TransVOD architecture is shown in Fig.~\ref{fig:framework_transvod}. It takes multiple frames of a video clip as inputs and outputs the detection results for the current frame. It contains four main components: Spatial Transformers for single frame object detection, extracting both object queries and compact features representation (memory for each frame), Temporal Deformable Transformer Encoder (TDTE) to fuse memory outputs from Spatial Transformers, Temporal Query Encoder (TQE) to link objects in each frame along the temporal dimension and Temporal Deformable Transformer Decoder (TDTD) to obtain final outputs for the current frame.

\vspace{-4mm}
\noindent
\textbf{Spatial Transformer.} We use Deformable DETR~\cite{zhu2020deformable} as our still image detector. In particular, to simplify complex designs in ~\cite{zhu2020deformable}, we \emph{do not} use multi-scale features in both Transformer encoders and decoders. We only use the last stage of the backbone as the input of the deformable Transformer. The modified detector includes Spatial Transformer Encoder (STE) and Spatial Transformer Decoder (STD), which encodes each frame $F$ (including Reference Frame and Current Frame) into two compact representations: spatial object query $Q$ and memory encoding $E$.
%(shown in blue arrow and black arrow in Figure~\ref{fig:framework}). 

\vspace{-4mm}
\noindent
\textbf{Temporal Deformable Transformer Encoder.}
The goal of TDTE is to encode the spatial-temporal feature representations and provide the location cues for the final decoder output. Since most adjacent features contain similar appearance information, directly using naive Transformer encoders~\cite{detr,Vaswani17attention} may bring much extra computation (much useless computation on object background). Deformable attention~\cite{zhu2020deformable} samples only partial information efficiently according to the learned offset field. Thus, we can link these memory encodings $E_{t}$ through this operation in a temporal dimension. The core idea of the temporal deformable attention modules is that we only attend to a small set of key sampling points around a reference efficiently.  Thus, TDTE receives the feature memories of the reference frame and the current frame as inputs, and outputs the enhanced current memory. The multi-head temporal deformable attention (TempDeformAttn) is as follows:
\begin{align}
    \text{TempDeformAttn}(z_q, \hat{p}_q, \{x^l\}_{l=1}^{L}) &=\sum_{m=1}^{M} W_m \big[\sum_{l=1}^{L} \sum_{k=1}^{K} A_{mlqk} \nonumber \\
    & x^l(\phi_l(\hat{p_q}) + \Delta p_{mlqk} )\big], \label{eq:def-attention}
\end{align}
where $m$ indexes the attention head, $l$ indexes the frame sampled from the same video clip, and $k$ indexes the sampling points, and $\Delta p_{mlqk}$ and $A_{mlqk}$ indicate the sampling offset and attention weights of the $k^\text{th}$ sampling point in the $l^\text{th}$ frame and the $m^\text{th}$ attention head, respectively.
$A_{mlqk}$ denotes the scalar attention weight in the range of $[0, 1]$, normalized by $\sum_{l=1}^{L} \sum_{k=1}^{K} A_{mlqk} = 1$. $\Delta p_{lmqk} \in R^2$ are of 2-d real numbers with unconstrained range. Since $p_q + \Delta p_{mlqk}$ is
fractional, we apply bilinear interpolation in \cite{dai2017deformable} for computing $x(p_q + \Delta p_{mlqk})$.
For each frame $l$, both $\Delta p_{mlqk}$ and $A_{mlqk}$ are calculated by feeding the query feature $z_q$ to a linear projection of $3MK$ channels, where the first $2MK$ channels encode the sampling offsets $\Delta p_{mlqk}$, and the remaining $MK$ channels are fed to a $\operatorname{Softmax}$ function to obtain the attention weights $A_{mlqk}$.
Here, we use normalized coordinates $\hat{p}_q \in [0, 1]^2$ for the clarity of scale formulation, in which  $(0, 0)$ and $(1, 1)$ indicate the top-left and the bottom-right image corners, respectively. $\phi_{l}(\hat{p}_q)$  re-scales the normalized coordinates $\hat{p}_q$ to the input feature map of  $l$-th frame. The multi-frame temporal deformable attention samples $LK$ points from $L$ feature maps instead of $K$ points from single-frame feature maps. There exist total $M$ attention heads in each TDTE layer.

\noindent
\textbf{Temporal Query Encoder.}
As mentioned in the previous part, learnable object queries can be regarded as the non-geometric anchors, which automatically learns the statistical features of the whole still image datasets during the training process. It means that the spatial object queries are not related to temporal contexts across different frames. Thus, we propose a \emph{simple yet effective} encoder to measure the interactions between the objects in the current frame and the objects in reference frames. 

Our key idea is to link these spatial object queries in each frame via a temporal Transformer, and thus learn the temporal contexts across different frames. We name our module Temporal Query Encoder (TQE). TQE takes all the spatial queries from reference frames to enhance the spatial output query of the current frame, and it outputs the temporal query for the current frame. Moreover, inspired from \cite{deng19rdn}, we design a coarse-to-fine spatial object query aggregation strategy to progressively schedule the interactions between the current object query and the reference object queries. The benefit of such a coarse-to-fine design is that we can reduce the computation cost to some extent. 

Specifically, we combine the spatial object query from all reference frames, denoted as $Q_{ref}$. Then, we perform the scoring and selection in a coarse-to-fine manner. In particular, we use an extra Feed Forward Networks (FFN) to predict the class logits, which are jointly trained with the spatial Transformers and the parameters are fixed when fine-tuning the temporal Transformers.
After that, we get the sigmoid value of that: $p=Sigmoid [ FFN(Q_{ref}) ]$. Then, we sort all the reference points by $p$ value and select the top-confident $k$ values from these reference points. The higher $p$ score means most likely objects and trained jointly with classification loss. The prediction head is only  trained for image object detection and is fixed for the training of video object detection. As most current DETR-like detectors~\cite{detr,zhu2020deformable} use the cascaded heads to refine detection results, we adopt a similar coarse-to-fine design to select less but precise object queries in the latter stages since most queries are not used and duplicated in the latter stages.

As shown in the blue part of Fig.~\ref{fig:framework_transvod}, TQE includes a self-attention layer, cross-attention, and FFN.
The temporal object queries are progressively refined and interacted with the spatial object queries extracted from different frames, calculating the co-attention between the reference queries and the query feature of the current frame. Note that the cross-attention plays the role of a cascade feature refiner which updates the output queries of each spatial Transformer iteratively. As such, TQE receives the object queries of the reference frames and the current frame as inputs and outputs the refined temporal object query of the current frame.

\begin{figure*}[h]
    \centering
    \includegraphics[width=1.00\textwidth]{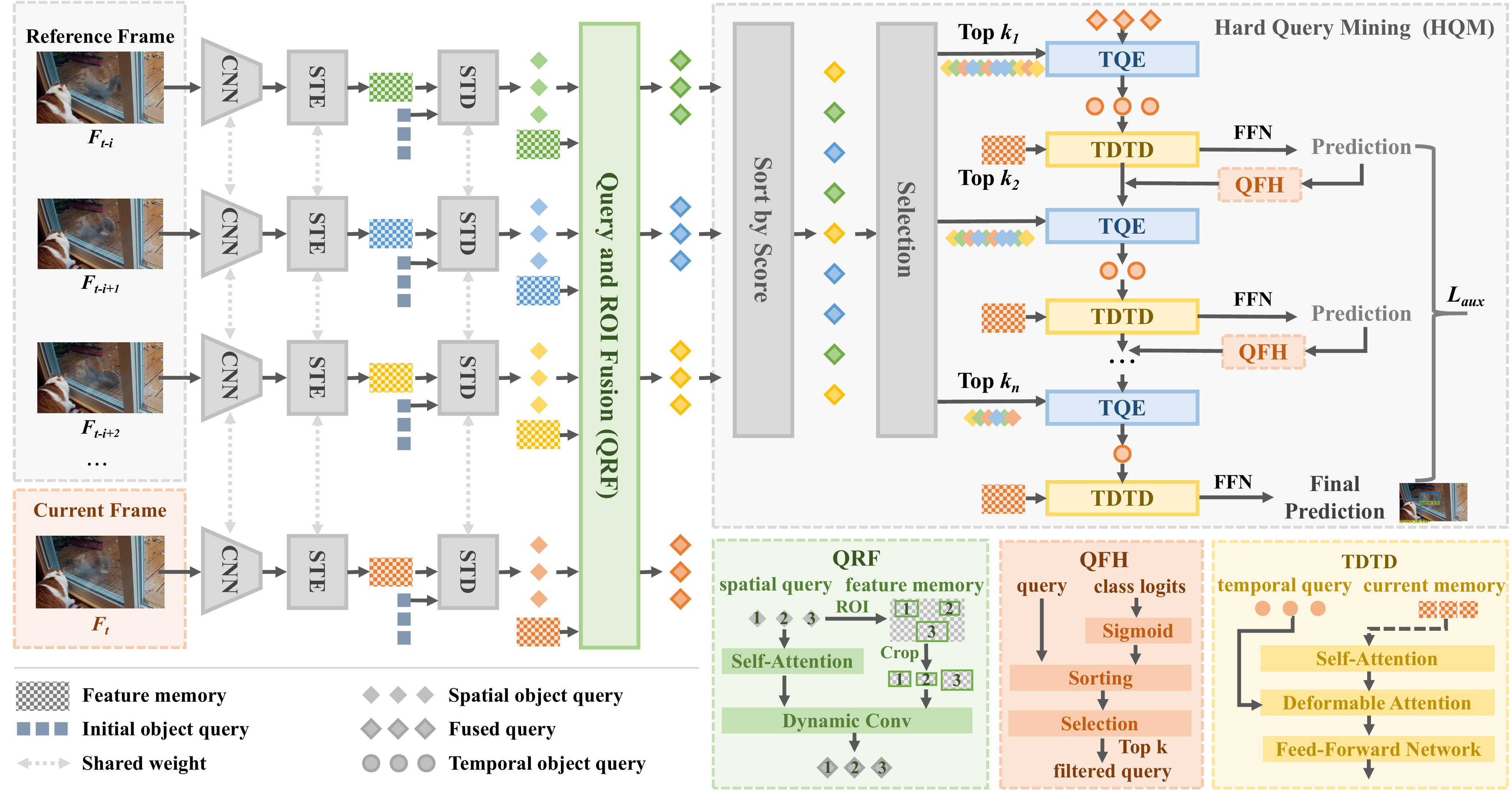}
    \caption{\small \textbf{The whole pipeline of TransVOD++.} Compared with the original TransVOD, TransVOD++ adds the Query and RoI Fusion (QRF) and Hard Query Mining (HQM) module. To avoid redundant spatial information in TDTE, we present QRF by fully injecting the object-level appearance information into each object query. Then, to dynamically reduce the query number and target number, we present HQM for mining the hardest query with multiple TDTD modules and multiple auxiliary TDTD losses. }
    \label{fig:framework_transvod_plus}
    \vspace{-5mm}
\end{figure*}

\noindent
\textbf{Temporal Deformable Transformer Decoder.}
This decoder aims to obtain the current frame output according to both outputs from TDTE (fused memory encodings) and TQE (temporal object queries). Given the aggregated feature memories $\hat{E}$  and the temporal queries $\hat{O_q}$, our Temporal Deformable Transformer Decoder (TDTD) performs co-attention between online queries and the temporal aggregated features.
The deformable co-attention~\cite{zhu2020deformable} of the temporal decoder layer is shown as follows:
\begin{align}
\text{DeformAttn}(z_q, p_q, x) = \sum_{m=1}^{M} W_m \big[\sum_{k=1}^{K} A_{mqk} \nonumber \\
\cdot W'_m x(p_q + \Delta  p_{mqk})\big],
\label{eq:single_deform_attn_fun}
\end{align}
where $m$ indexes the attention head, $k$ indexes the sampled keys, and $K$ is the total number of the sampled keys ($K \ll HW$).
$p_{mqk}$ and $A_{mqk}$ indicate the sampling offset and attention weight of the $k^\text{th}$ sampling point in the $m^\text{th}$ attention head, respectively.
The attention weight $A_{mqk} \in [0, 1]$, normalized by $\sum_{k=1}^{K} A_{mqk} = 1$. $\Delta p_{mqk} \in R^2$ are of 2-d real numbers with unconstrained range. Due to the fact that $p_q + \Delta p_{mqk}$ is fractional, we also adopt bilinear interpolation in computing $x(p_q + \Delta p_{mqk})$ following \cite{dai2017deformable}. 
Both $\Delta p_{mqk}$ and $A_{mqk}$ are obtained via linear projection over the query feature $z_q$. In our implementation, the query feature $z_q$ is fed to a linear projection operator. The output of TDTD is sent to one feed-forward network (FFN) for the final classification and box regression as the detection results of the current frame.

\vspace{-4mm}
\subsection{TransVOD++}
\label{sec:TransVOD++}
Compared with previous work, despite TransVOD simplifying the pipeline of VOD, it has several limitations. Firstly, it contains heavy computation costs in TDTE. Secondly, the performance of TransVOD is still limited. To solve these problems, we present TransVOD++ which contains the following improvements including Query and RoI Fusion (QRF), Hard Query Mining (HQM), and a strong backbone. The pipeline is shown in Fig~\ref{fig:framework_transvod_plus}.

\noindent
\textbf{Query and RoI Fusion.} Previous works~\cite{chen2020memory,hu18relationnet} show that region features are useful and contain precise appearance information for temporal fusion. Our motivation is to replace TDTE with features in region of interest (RoI) via the proxy strategy where each RoI feature is injected into each query, thus utilizing the object-level appearance information to enhance the object query.

Specifically, given the detection boxes from spatial Transformers, we get the region of interest (RoI) of each frame in a video clip. Then, according to those RoIs and the feature from the STE, we calculate the RoI feature $E^{RoI}_{cur}$ and $E^{RoI}_{ref}$ of the current frame and the reference frames, respectively. Next, the cropped RoI features are used to weigh each query via the transformation of MLP, as shown in the green part of Fig.~\ref{fig:framework_transvod_plus}.
The current RoI feature $E^{RoI}_{cur}$ is aggregated onto the object query of the current frame to generate the enhanced current query feature $\hat{Q}_{cur}$, where feature aggregation is conducted through dynamic convolutions.  
\begin{equation}
\hat{Q}^{j+1}_{cur} =\left\{\begin{array}{l}
\text{QRF}(Q^{j}_{cur}, E^{RoI}_{cur}), \text { if } j=1  \\
\text{QRF}(\hat{Q}^{j}_{cur}, E^{RoI}_{cur}), \text { otherwise }
\end{array}\right.
\end{equation}
where $Q^{j}_{cur}$ is the spatial object query of the current frame before the $j_{th}$ temporal query encoder (TQE), and $\hat{Q}^{j}_{cur}$ denotes the temporal object query before the $j_{th}$ TQE module.
Similarly, for each reference frame, the reference RoI features $E^{RoI}_{cur}$ of the $i_{th}$ frame are fused with the reference query of the $i_{th}$ frame via QRF.

The details of the QRF module are described as follows: given the object query and RoI feature memory, we first feed the object query to a multi-head self-attention layer to reason about the relations between objects. Then, each RoI feature will interact with the corresponding object query to filter out ineffective bins and outputs the final object query. Inspired from \cite{sun2021sparse}, we carry out two consecutive $1 \times 1$ convolutions with ReLU activation function for light design. 
The $k_{th}$ object query generates dynamic parameters of these two convolutions for the corresponding $k_{th}$ RoI feature via a linear projection.
Finally, the aggregated reference queries $\hat{Q}^{j}_{ref}$ are used to enhance the aggregated current query $\hat{Q}^{j}_{cur}$ via a TQE, thus learning the temporal contexts across different frames, which is described as: $\hat{Q}^{j}_{cur} = \text{TQE}(\hat{Q}^{j}_{cur}, \hat{Q}^{j}_{ref})$

\noindent
\textbf{Hard Query Mining.}
% by qianyu 
Considering that both the spatial object queries and temporal object queries contain much redundant information across the dataset, for example, 300 queries reflect the temporal appearance distributions of 30 categories, and those queries need to match more than 300 ground truths during the training procedure, and there is no need to maintain so many object queries/targets in both the spatial and temporal dimension. As such, we are motivated to dynamically reduce the redundancy of query number and target number in the training of temporal Transformers, and meanwhile, we mine the hardest query in both the current frame and the reference frames.

\begin{figure*}[h]
    \centering
    \includegraphics[width=1.00\textwidth]{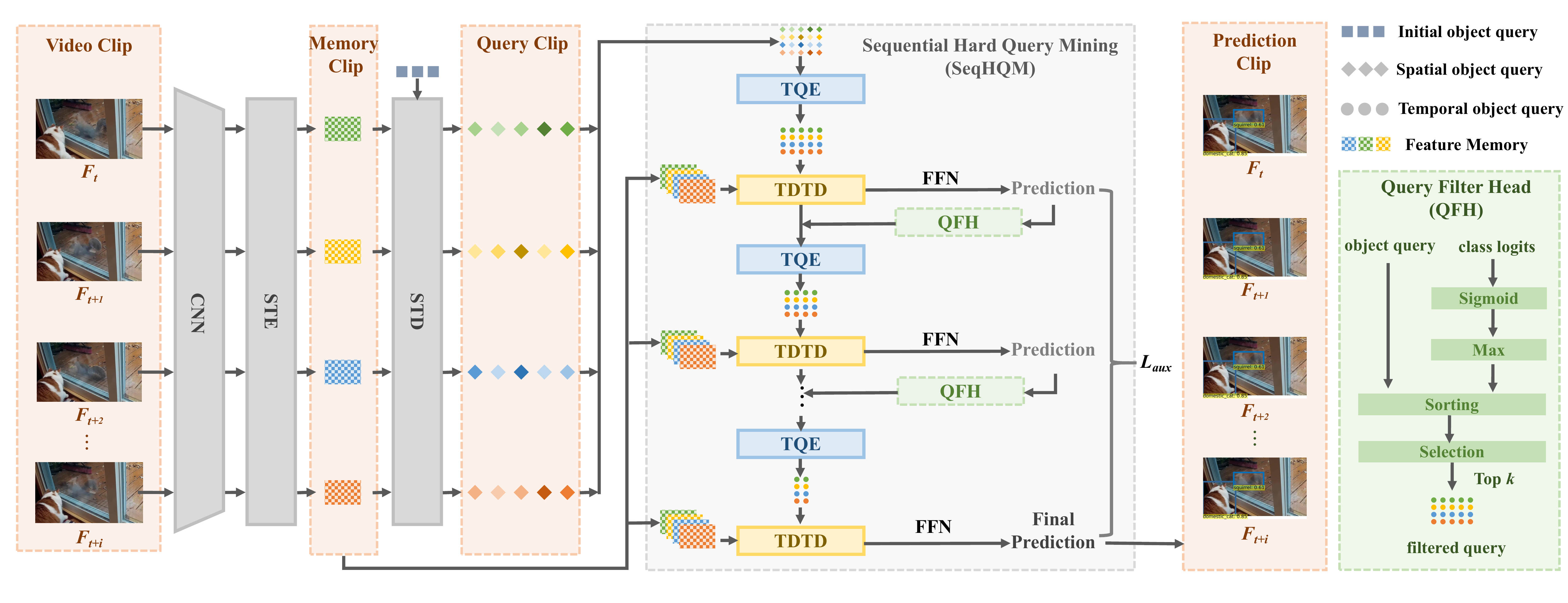} 
    \vspace{-6mm}
    \caption{\small \textbf{The whole pipeline of TransVOD Lite.} Compared with TransVOD and TransVOD++, TransVOD Lite aims at real-time video object detection. It takes multiple frames as inputs and outputs all the results simultaneously. We propose Sequential Hard Query Mining (SeqHQM) to mine the hardest query in a video clip for selectively reducing the redundancy of sequential object queries and targets in the training of temporal Transformers via Query Filter Head (QFH) and auxiliary TDTD losses $\mathcal{L}_{aux}$. }
    \vspace{-5mm}
    \label{fig:framework_tranvod_lite}
\end{figure*}

Concretely, given the spatial object query $Q_{ref}$ of the reference frames, we fed it into a Query Filter Head (QFH), which filters the redundant object query and select the most confident ones to reduce the computation redundancy. Specifically, $Q_{ref}$ are fed forward to the class embedding layer of the spatial Transformer, \emph{i.e.,} a linear classification layer with sigmoid activation, to generate class logits. Then, those reference queries are concatenated in the dimension of the query number. Next, according to the probability of the reference logits, we sort and then select the top $k$ confident query, which is illustrated in the salmon part of Fig.~\ref{fig:framework_transvod_plus}. Inherited from TransVOD, we adopt the coarse-to-fine query aggregation strategy to progressively model the relationships between the current query and the reference queries via TQE module.

\textit{The differences between TransVOD++ and TransVOD lie in several aspects.}
Firstly, in contrast to TransVOD that only selects the reference query, our TransVOD++ selects not only the reference query but also the current query. Both of them are treated differently in a coarse-to-fine manner, thus reducing the computation cost in the temporal Transformer. Secondly, compared to TransVOD, we add a Temporal Defomrable Transformer Decoder (TDTD) after each TQE module and supervise the object query with different query numbers via an \textbf{auxiliary TDTD loss}, denoted as $\mathcal{L}_{aux}$. We find it helpful to use auxiliary TDTD losses $\mathcal{L}_{aux}$ in temporal Transformer during training, especially to help the model output the correct number of objects of each class. We add prediction FFNs and Hungarian loss after each TDTD module. All prediction FFNs share their parameters.

\noindent
\textbf{Strong Backbone.} We further adopt Swin Transformer as the strong backbone network. However, Swin Transformer~\cite{liu2021swin} generates multi-scale features adopted with FPN-like framework~\cite{he17maskrcnn} which is not suitable for our TransVOD framework. We propose a simple yet effective solution via fusing multi-scale features into one scale where we directly add multi-scale features into one scale.

\subsection{TransVOD Lite}
\label{sec:TransVOD_lite}
Despite TransVOD and TransVOD++ make the VOD pipeline much simpler, the inference time is still limited due to multiple frame query fusing. As mentioned in Section~\ref{sec:related}, the inference time is critical for real-world applications. To embrace the advantage of modeling sequence data in Transformer~\cite{vaswani2017attention,wang2020end}, we present TransVOD Lite where it takes multi frames as inputs and output detection results of all frames directly, as shown in Fig.~\ref{fig:framework_tranvod_lite}. 

\noindent
\textbf{Direct Multiple Frame Predictions.} In TransVOD Lite, we abandon the feature aggregation paradigm, which requires much more computation costs in terms of time and memory space. Instead, a sequence of video clips is fed as input and output a sequence of results. 
As shown in Fig.~\ref{fig:framework_tranvod_lite}, TransVOD Lite inherits the Hard Query Mining from the TransVOD++ and spatial-temporal transformer design in TransVOD including TQE, and TDTD. The main difference is that TransVOD Lite directly outputs the multiple frame prediction with a hyper-parameter $T_w$ which is the temporal window size of the input clip or the number of the input frames. When $T_w$ is larger, the inference speed is faster while the memory is increased. In this way, we can fully use the memory of GPU to speed up the inference time. We provide detailed experiments on the effect of choosing $T_w$ in the experiment part.

\noindent
\textbf{Sequential Hard Query Mining.} 
Different from TransVOD and TransVOD++, we do not need to discriminate whether an object query is the reference query or the current query for filtering, all object queries in the whole sequence are equally selected in a coarse-to-fine manner, thus increasing the speed, \emph{e.g.,} FPS, to $T_w$ times in temporal Transformer than original TransVOD, where $T_w$ denotes the temporal window size in a given clip.
We name our method ``sequential hard query mining'' (SeqHQM).
For example, $T_w=12$ means the input frames are 12 in the video clip, and then we need to generate the results of those 12 frames, if each frame has 300 object queries, there are 3600 object queries in total. There is no doubt that there exists large redundant information of those large number of object queries, and it is necessary to dynamically reduce the computation costs to boost the inference speed, as well as achieve good results in modeling the temporal motion.

We then describe SeqHQM in detail. Specifically, a sequence of spatial object query $Q_{seq}$, is fed forwarded into a Query Filter Head (QFH) to select the most credible object queries. The number of object queries and targets is dynamically decreasing to reduce the computation redundancy. For TransVOD Lite, we implement the QFH differently before the $k_{th}$ TQE module. If $k=1$, we use the class embedding layer of the spatial Transformer to generate class logits and go through a sigmoid activation function, which is similar as QFH in TransVOD ++. If $k>1$, the class logits are generated through the learnable temporal class embedding layer then with a sigmoid activation function. Next, we compute the maximum probability and select the top $k$ confident query by sorting and selection in a coarse-to-fine manner, which is illustrated in the green part of Fig.~\ref{fig:framework_tranvod_lite}.
Similar to TransVOD++, we add a TDTD after each TQE module and supervise the object query with different query numbers via an \textbf{auxiliary TDTD loss}, denoted as $\mathcal{L}_{aux}$.  $\mathcal{L}_{aux}$ is essential to help the model output the correct number of objects of each class. We add prediction FFNs and Hungarian loss after each TDTD module. All prediction FFNs share their parameters.

\subsection{Loss Functions and Inference } 
\label{sec:loss_inference}
\noindent
\textbf{Loss functions.}
Original DETR~\cite{detr} avoids post-processing and adopts a one-to-one label assignment rule. Following~\cite{stewart2016end, detr, zhu2020deformable}, we match predictions from STD/TDTD with ground truth by Hungarian algorithm~\cite{kuhn1955hungarian} and thus the entire training process of spatial Transformer is the same as original DETR. 
The temporal Transformer uses similar loss functions given the box and class prediction output by the FFN. The matching cost is defined as the loss function. Following \cite{detr, zhu2020deformable, sun2021sparse}, the loss function is:
\begin{align}
    \label{eq:loss}
    \mathcal{L}_{aux} = \sum_{j=1}^{J} \big[ \lambda_{cls}\cdot\mathcal{L}_{\mathit{cls}}+\lambda_{L1} \cdot \mathcal{L}_{\mathit{L1}}+\lambda_{giou}\cdot\mathcal{L}_{\mathit{giou}} \big ],
\end{align}
where $J$ denotes the total number of TDTD modules in the temporal Transformers, where $J$ = 1 for TransVOD and $J$ = 3 for TransVOD++ and TransVOD Lite in all experiments. 
$\mathcal{L}_{\mathit{cls}}$ represents focal loss~\cite{FocalLoss} for classification. $\mathcal{L}_{\mathit{L1}}$ and $\mathcal{L}_{\mathit{giou}}$ represent L1 loss and generalized IoU loss~\cite{GIoU} in for localization. $\lambda_{cls}$, $\lambda_{L1}$ and $\lambda_{giou}$ are coefficients of them. We balance these loss functions following the same setting in ~\cite{zhu2020deformable}.
For TransVOD Lite, we apply such a loss function for all input frames.

% \vspace{-4mm}
\noindent
\textbf{Inference for TransVOD Lite.}
In TransVOD Lite, the window size of a given video is defined as $T_w$ and the interval between the two adjacent frames within one clip is denoted as $I_w$, respectively.  Given a video $V = \{F_1, F_2, \cdots, F_N\}$, we first expand the video size to the integer multiples of $T_w$ as: $\hat{N} = \lceil \frac{N}{T_w}  \rceil T_w$. Then, for each expanded video, we divide the video into two parts and adopt different sampling strategies for these two parts. 

As for the first part, the clip is normal where the interval of different frames is $I_w$.
The index of the first frame in each video clip is $S = T_wI_w i + j $, where $i \in \{ 0,1, \cdots, K -1 \}$, $j \in \{1, \cdots, I_w -1 \}$ , $K = \lfloor \frac{\hat{N}}{T_wI_w}  \rfloor $. We feed the normal clip sequentially with window size $T_w$ and interval size $I_w$ into the model. For the second part, the frames are not divisible by $T_w I_w$. The index of the first frame is the clip is $T_w k +1$. There are $\hat{N} - TW k$ frames in this clip. Those frames are randomly divided into $\frac{\hat{N}}{T_w} - KI_w$ video clips, with the size of each clip as $T_w$.  

Besides, we introduce another sampling strategy using random shuffling.
We find that if we first randomly shuffle $\hat{v}$ and split it to $\frac{\hat{N}}{T_w}$ clips, our model could model the temporal motions better due to the large view of the video. The empirical evidence perceived by the human visual system illustrates that when people are not certain about the identity of an object, they would seek to find a distinct object from other
frames that share high semantic similarity with the current object and assign them together. Regarding that Transformers are effective in modeling long-range dependencies, if we randomly shuffle the video, we could increase the data diversity and fully utilize the global information of the video. The effectiveness of both strategies is demonstrated in Sec.~\ref{sec:ablation_transvod_lite}.

\section{Experiment}
% \noindent
% \textbf{Overview.} In this section, we first introduce the experimental setup, \emph{e.g.,} dataset and implementation details, of TransVOD. Then, we compare our methods with the state-of-the-art VOD methods in various settings. Next, we perform several ablation studies to validate the effectiveness of all three models. Finally, we provide visualization and analysis to reveal more insights.

\subsection{Experimental Setup}
% \subsubsection{Datasets}
\noindent 
\textbf{Datasets:} We empirically conduct experiments on the ImageNet VID dataset~\cite{russakovsky2015imagenet} which is a large-scale benchmark for video object detection. It contains 3862 training videos and 555 validation videos
with annotated bounding boxes of 30 classes. Since the ground truth of the official
testing set is not publicly available, following common VOD protocols~\cite{zhu17fgfa, wang18manet, deng19rdn, wu19selsa}, we train our models using a combination of ImageNet VID and DET datasets~\cite{russakovsky2015imagenet} and measure the performance on the validation set using mean average precision (mAP) metric.

% \subsubsection{Implementation details.}
\noindent
\textbf{Network architectures:} In this work, we use Deformable DETR~\cite{zhu2020deformable} as the image detector, and following~\cite{yao2020video,liu2019looking,qian2020adaptive}, the detector is pre-trained on the COCO dataset~\cite{COCO_dataset}. 
Following the
widely used implementation protocols in previous works~\cite{zhu17fgfa, wang18manet, deng19rdn, wu19selsa}, We use ResNet-50~\cite{he16res} and ResNet-101~\cite{he16res} as the network backbone. Besides, we also adopt Swin Transformer~\cite{liu2021swin} as the backbone 
for better performances, which uses the same hyper-parameters as ResNet backbone. Note that we do not use the  multi-scale features of the FPN-like network and fuse the multi-scale features by adding into the largest scale.
All these backbones are pre-trained on ImageNet~\cite{deng2009imagenet} dataset.
More implementation details could be referred in our \href{https://github.com/SJTU-LuHe/TransVOD}{\textit{code link}}.

\noindent 
\textbf{Training details:} 
Following Deformable DETR~\cite{zhu2020deformable}, we use the  AdamW~\cite{loshchilov2017decoupled} optimizer, the initial learning rate is $2\times 10^{-4}$ for Transformers, and $2\times 10^{-5}$ for the backbone, and weight decay is $10^{-4}$. All Transformer weights are initialized with Xavier~\cite{glorot2010understanding}. 
The number of initial object queries is set as 300 for ResNet~\cite{he16res} and 100 for Swin Transformer~\cite{liu2021swin}.
During the training, the batch size is 1, and
the number of reference frames is 14 for TransVOD and TransVOD++ in all experiments.
Following the sampling strategy in MMTracking~\cite{mmtrack2020}, we adopt the bilateral uniform sampling for reference frames, which means reference images are randomly
sampled from the two sides of the nearby frames of the current frame.
For TransVOD lite, the total frames of the video clip are sequentially fed into the model. In all experiments, we use the same data augmentation as MEGA~\cite{chen2020memory}, including random horizontal flip, randomly resizing. We train the model for 14 epochs in an end-to-end manner. For better convergence, we first train the spatial Transformers for 7 epochs and then fine-tune the temporal Transformers for another 7 epochs. In the fine-tuning process, we freeze the parameters of spatial Transformers and only optimize the temporal Transformers.

\noindent 
\textbf{Inference details:} 
The inference runtime (FPS) of Table~\ref{table:transvod_lite_stoa} is calculated on a single V100 GPU card. During the inference phase, the batch size is 1 and we sample the reference frame with a fixed frame stride for TransVOD and TransVOD++. As mentioned in Sec~\ref{sec:loss_inference}, we sample the frames via random shuffling for TransVOD Lite.
During the inference phase, we use the same data augmentation as MEGA~\cite{chen2020memory} for image resizing such that the shortest side of the image is at least 600 while the longest is at most 1000. 
Note that we do not need any sophisticated post-processing method, which largely simplifies the pipeline of VOD. 
% % %For saving the computation cost, we 
% We both the spatial Transformer encoder and the spatial Transformer decoder's weight are fixed to save the computation cost.
% \noindent \textbf{Training.} 
% \noindent \textbf{Inference.}

\subsection{Main Results}

We first compare our proposed TransVOD and TransVOD++ using ResNet-50 backbone in Table~\ref{table:mainresult_r50}. Then we present the detailed results with the previous state-of-the-art VOD methods in Table~\ref{table:mainresult_r101}. Finally, we compare the real-time VOD models in Table~\ref{table:transvod_lite_stoa}.

\noindent
\textbf{Results using ResNet-50 backbone.}
Table~\ref{table:mainresult_r50} shows the comparison results with the state-of-the-art VOD methods with ResNet-50 backbone.
For a fair comparison, we also report the performance of existing VOD methods that use the COCO-pretraining model.
Despite the use of COCO-pretraining weights boosts the mAP of existing VOD methods, our proposed TransVOD still achieves superior performance against the state-of-the-art methods by a large margin. 
In particular, TransVOD achieves 79.9$\%$ with ResNet-50, which makes 1.3\%$\sim$2.6$\%$ absolute improvements over the best competitor MEGA~\cite{chen2020memory}. Moreover, our proposed TransVOD++ further improves the original TransVOD by 0.6$\%$, achieving 80.5$\%$ on the ImageNet VID validation set.

% \vspace{2mm}
\noindent
\textbf{Results with stronger backbone.} We further report stronger backbone results to compare with the state-of-the-art methods in Table~\ref{table:mainresult_r101}. When equipped with a stronger backbone ResNet-101, the mAP of our TransVOD++ is further boosted up to 82.0\%, which outperforms most state-of-the-art VOD methods~\cite{zhu17dff,feichtenhofer17dt,zhu17fgfa,wang18manet,zhu18hp}. Specifically, our model is remarkably better than FGFA~\cite{zhu17fgfa} (76.3$\%$ mAP) and MANet~\cite{wang18manet} (78.1$\%$ mAP), which both aggregate features based on optical flow estimation, and the mAP improvements are +5.6$\%$ mAP and +3.8$\%$ mAP respectively. When compared with some relation-based methods
(LRTRN~\cite{shvets19lltr} (81.0$\%$ mAP), RDN~\cite{deng19rdn} (81.8$\%$ mAP), SELSA~\cite{wu19selsa}  (80.3 $\%$ mAP)), our method also shows its superiority in case of detection precision.
Moreover, our proposed method boosts the strong baseline \emph{i.e.,} deformable DETR~\cite{zhu2020deformable} by a significant margin (\textbf{3\%$\sim$4\% mAP}). After adopting Swin Base (SwinB) as the backbone, our TransVOD++ achieve \textbf{90.0$\%$ mAP} and it outperforms previous works by a large margin (about 4 \%$\sim$ 5\% mAP), which further demonstrate its effectiveness. 

\begin{table}[t!]
\caption{
Comparison with state-of-the-art methods on ImageNet VID with Res50 backbone. $^{\dag}$ means COCO pre-training.}
\footnotesize
\vspace{-3mm}
\begin{center}
\begin{tabular}{c|c|c}
\toprule
Methods& Base Detector  & mAP (\%) \\
% \midrule
% DFF~\cite{zhu17dff}  & R-FCN & 10  &70.4\\
% FGFA~\cite{zhu17fgfa}  & R-FCN & 21  &74.0\\
\midrule
Single Frame Baseline~\cite{ren2016faster} &  Faster-RCNN & 71.8 \\
DFF~\cite{zhu17dff}  & Faster-RCNN &70.4\\
FGFA~\cite{zhu17fgfa}  & Faster-RCNN &74.0\\
RDN~\cite{deng19rdn} &  Faster-RCNN & 76.7\\
MEGA~\cite{chen2020memory}  & Faster-RCNN & 77.3\\
\midrule
Single Frame Baseline$^{\dag}$~\cite{ren2016faster} &  Faster-RCNN$^{\dag}$ & 72.7\\
DFF$^{\dag}$~\cite{zhu17dff}  & Faster-RCNN$^{\dag}$ &71.6\\
FGFA$^{\dag}$~\cite{zhu17fgfa}  & Faster-RCNN$^{\dag}$ &75.1\\
RDN$^{\dag}$~\cite{deng19rdn} &  Faster-RCNN$^{\dag}$ & 77.6\\
MEGA$^{\dag}$~\cite{chen2020memory}  & Faster-RCNN$^{\dag}$ & 78.3\\
\midrule
Single Frame Baseline~\cite{zhu2020deformable}  & Deformable DETR & 76.0 \\
TransVOD &  Deformable DETR  & \textbf{79.9} \\
TransVOD++ &  Deformable DETR  & \textbf{80.5} \\
\bottomrule
\end{tabular}
\end{center}
\label{table:mainresult_r50}
\vspace{-1mm}
\end{table}%

\begin{table}[t]\centering
\caption{
Performance comparison with state-of-the-art \textit{real-time VOD methods} on ImageNet VID validation set. In terms of both accuracy and speed, 
Our method outperforms most of them and has fewer parameters than existing models. 
}
\scalebox{0.92}{%
\begin{tabular}{lcccc}
\toprule
Model   &  mAP (\%) & \thead{Runtime\\ (FPS)}  & \thead{\#Params\\ (M)}  & Backbone \\
\midrule
DFF \cite{zhu17dff}  & 73.1 & 20.25 &97.8  & Res101  \\
D \&T  \cite{feichtenhofer17dt} & 75.8 & 7.8 &-     & Res101 \\
LWDN \cite{jiang2019video}  & 76.3 & 20  &77.5 & Res101 \\
OGEMNet\cite{deng2019ogemn} & 76.8  & 14.9 &- & Res101 \\
THP \cite{zhu18hp} & 78.6 & 13.0 &-& Res101+DCN \\
RDN  \cite{deng19rdn}  & 81.8  &  10.6 &- & Res101 \\
SELSA \cite{wu19selsa}  & 80.3  & 7.2  &-  & Res101   \\
LRTR \cite{shvets19lltr}  & 80.6  & 10 &- & Res101 \\
PSLA \cite{guo2019progressive}  &77.1 & 18.7 & 63.7 & Res101   \\
PSLA \cite{guo2019progressive}  & 80.0 & 13.3 & 72.2  & Res101+DCN  \\
LSTS \cite{jiang2020learning}  &77.2 & 23.0 & 64.5& Res101   \\
LSTS \cite{jiang2020learning} & 80.1 & 21.2 & 65.5  & Res101+DCN  \\
% \midrule
\midrule
TransVOD Lite &80.5 & \textbf{32.3} &74.2 & Res101   \\
TransVOD Lite  &83.7 & 29.6 &\textbf{46.9} & SwinT   \\
TransVOD Lite  &85.8 & 22.2 &68.3 & SwinS   \\
TransVOD Lite  &\textbf{90.1} & 14.9 &106.3 & SwinB   \\
%\hline
\bottomrule
\end{tabular}
}
\label{table:transvod_lite_stoa}
\end{table}

% Table generated by Excel2LaTeX from sheet 'cross'
\begin{table}[t!]
\caption{Comparison with the state-of-the-art VOD methods on ImageNet VID. Most VOD methods use ResNet 101 as the backbone. $^{\star}$ denotes using Swin-Base as backbone.}
\footnotesize
\vspace{-3mm}
\begin{center}
\begin{tabular}{c|c|c}
\toprule
Methods & Base Detector &  mAP(\%) \\
\midrule
Single Frame Baseline~\cite{dai2016r}  & R-FCN & 73.6 \\ % -+
DFF~\cite{zhu17dff}  & R-FCN &73.0\\                % -+
AdaScale~\cite{feichtenhofer17dt}  & R-FCN &75.5 \\  % -+
D$\&$T~\cite{chin2019adascale}  & R-FCN &75.8 \\  % -+
FGFA~\cite{zhu17fgfa}& R-FCN &76.3\\             % -+
LWDN~\cite{jiang2019video} & R-FCN &76.3\\
IFF-Net~\cite{jin2022feature} & R-FCN &77.1\\
SCNet~\cite{wang2020scnet} & R-FCN &77.9\\
AFA~\cite{qian2020adaptive} & R-FCN &77.9\\
% MANet~\cite{wang18manet}& R-FCN  &78.1\\         % -+
THP~\cite{zhu18hp} & R-FCN &78.6\\           % -+
STSN~\cite{bertasius18stsn}  & R-FCN  &78.9\\   % -
PSLA~\cite{guo2019progressive}  & R-FCN  & 80.0 \\% -+
OGEMN~\cite{deng2019ogemn}  & R-FCN  & 80.0 \\ % 
 STMN~\cite{xiao18stmn}  & R-FCN  & 80.5 \\ % -
TCENet~\cite{he2020temporal}  & R-FCN  & 80.3 \\
MAMBA~\cite{sun2021mamba}  & R-FCN  & 80.8 \\
\midrule
Single Frame Baseline~\cite{ren2016faster} & Faster RCNN & 76.7 \\%+
ST-Lattice~\cite{chen2018optimizing} &  Faster RCNN &79.0\\%+
BFAN~\cite{wu2020bfan} &  Faster RCNN &79.1\\%+
STCA~\cite{luo2019object}   &  Faster RCNN & 80.3 \\
SELSA~\cite{wu19selsa}& Faster RCNN & 80.3\\% -+
MINet~\cite{MINet}& Faster RCNN & 80.6\\
LRTR~\cite{shvets19lltr} & Faster RCNN  & 81.0 \\% -+
RDN~\cite{deng19rdn}   & Faster RCNN & 81.8\\% -
TROI~\cite{gong2021temporal}   & Faster RCNN & 82.0\\% -
MEGA~\cite{chen2020memory} & Faster RCNN  & 82.9\\% -
HVRNet~\cite{han2020mining} & Faster RCNN  & 83.2\\
TF-Blender~\cite{Cui_2021_ICCV} & Faster RCNN  & 83.8\\
DSFNet~\cite{lin2020dual} & Faster RCNN  & 84.1\\
MAMBA~\cite{sun2021mamba} & Faster RCNN  & 84.6\\
EBFA~\cite{han2020exploiting} & Faster RCNN  & 84.8\\
CFA-Net~\cite{han2021cfanet} & Faster RCNN  & 85.0\\
% CSMN~\cite{han2021context} & Faster RCNN  & 85.2\\
\midrule
Single Frame Baseline~\cite{zhou2019objects}  & CenterNet  &73.6\\ %
CHP~\cite{xu2020centernet} &  CenterNet &76.7\\ % 
\midrule
Single Frame Baseline~\cite{zhu2020deformable} &  Deformable DETR & 78.3 \\
TransVOD Lite &Deformable DETR & 80.5 \\
TransVOD++ &Deformable DETR & 82.0 \\
TransVOD++$^{\star}$ &Deformable DETR & \textbf{90.0} \\
\bottomrule
\end{tabular}
\end{center}
\label{table:mainresult_r101}
\vspace{-5mm}
\end{table}%

\begin{table*}[h!]
 \footnotesize
 \centering
 \caption{Ablation studies of TransVOD on ImageNet VID using ResNet 50 as the backbone.}
    \subfloat[
    Effect of each component in TransVOD. TDTE: Temporal Deformable Transformer Encoder. TQE: Temporal Query Encoder. TDTD: Temporal Deformable Transformer Decoder. $\dag$ means the results using the Swin-base backbone.
    ]{
     \begin{tabularx}{0.52\textwidth}{c c c c |c c c } 
              \toprule[0.18em]
      Single Frame Baseline & TDTE & TQE & TDTD & mAP (\%) & mAP$^{\dag}$ (\%) \\
            \midrule[0.10em]
      \checkmark  &  &  & & 76.0 & 88.3\\
     \checkmark  & \checkmark & & \checkmark & 77.1 & 88.8\\
    \checkmark  &  & \checkmark &  & 78.9 & 89.3\\
    \checkmark  &   & \checkmark & \checkmark & 79.3 & 89.6\\
    \checkmark  & \checkmark & \checkmark & \checkmark  & 79.9 & 89.6\\
    \bottomrule[0.15em]
     \end{tabularx}
    }
    \hfill
    \subfloat[Ablation studies on the number of of top k spatial object query in three temporal deformable transformer encoder (TDTE) layers. Our coarse-to-fine (C2F) temporal query aggregation strategy has better results.]{
     \begin{tabularx}{0.45\textwidth}{c|ccccccc} 
     \toprule[0.15em]
    k$_j$ & I & II &III & IV  &V & VI & VII \\
    \midrule[0.15em]
    k1& 30 &30 &30 & 50  &50 & 80 & 80 \\
    k2& 20 &20 &30 & 30  &50 & 50 & 80  \\
    k3& 10 &20 &30 & 20  &50 & 20 & 80 \\
    \midrule[0.15em]
    mAP(\%) & 79.7 & 79.6 & 79.3 & 79.6 &  79.5 & 79.9 & 79.7 \\
    \bottomrule[0.15em]
     \end{tabularx}
    } \hfill
\label{tab:transVOD_ablation_1}
\end{table*}

\begin{table}[ht]
    \footnotesize
    \begin{center}
    \addtolength{\tabcolsep}{1.5pt}
    \caption{Ablation studies on TransVOD: number of encoder layers $N_{TDTE}$ in TDTE, number of encoder layers $N_{TQE}$ in TQE, number of decoder layers  $N_{TDTD}$ in TDTD and top $k$ spatial query in TQE with one decoder layer.}
    \label{tab:transVOD_ablation_2}
    \begin{tabular}{c|cccccc}
\Xhline{1.0pt}
\multicolumn{7}{c}{(a) {Number of encoder layers $N_{TDTE}$ in TDTE.}}\\
\hline
$N_{TDTE}$ &  0 & 1 & 2 & 3 & 4\\
%\hline
mAP(\%) & 77.0 & 77.7 & 77.6 & 77.8 & 77.7 \\
\Xhline{1.0pt}
\multicolumn{7}{c}{(b) {number of encoder layers $N_{TQE}$ in TQE}}\\
\hline
$N_{TQE}$ & 1 & 2 & 3  & 4 & 5 & 6\\
%\hline
mAP(\%)  & 78.8  & 79.4  & 79.6 & 79.6  & 79.7 &  79.7 \\
\Xhline{1.0pt}
\multicolumn{7}{c}{(c) {number of decoder layers  $N_{TDTD}$ in TDTD.}}\\
\hline
$N_{TDTD}$ & 1 & 2  & 3 & 4 & 5 & 6\\
%\hline
mAP (\%)  & 78.2 & 77.7 & 77.1 & 76.2 & 74.8 & 72.3 \\
\Xhline{1.0pt}
\multicolumn{7}{c}{(d) {top $k$ spatial query in TQE with one decoder layer.}}\\
\hline
$k$ & 25 & 50 & 100  & 200 & 300\\
mAP(\%) & 78.0 & 78.1 & 78.3 & 77.9 &  77.7 \\
\Xhline{1.0pt}
\multicolumn{7}{c}{(d) {Number of reference frames $N_{ref}$.}}\\
\hline
$N_{ref}$  & 2 & 4 & 8  & 10 & 12 & 14\\
mAP(\%) & 77.7 & 78.3  & 79.0  & 79.1 & 79.0& 79.3 \\
\Xhline{1.0pt}
    \end{tabular}
    \end{center}
\end{table}
%\vspace{2mm}

\noindent
\textbf{Results using TransVOD Lite}
In Table~\ref{table:transvod_lite_stoa}, we report the results of our TransVOD Lite and compare it with previous real-time VOD models. As shown in that table, using the ResNet-101 backbone, our method achieves the best speed and accuracy trade-off. After adopting Swin-Tiny as the backbone, our TranVOD Lite achieves \textbf{83.7 $\%$ mAP} while running at nearly 30 FPS. Our best TransVOD Lite model with a Swin base backbone can achieve 90.1 $\%$ mAP while running at around 15 FPS. Furthermore, the parameter count (46.9M) is fewer than other video object detectors (\emph{e.g.,} around 100M in ~\cite{zhu17dff}), which also indicates that our method is more friendly for mobile devices.

\subsection{Ablation Study and Analysis}

%To demonstrate the effect of key components in our proposed method, we conduct extensive experiments to study how they contribute to the final performance using ResNet-50 as backbone.  \\
\noindent
\textbf{Overview.} In this section, we demonstrate the effect of key components in our proposed methods including TransVOD, TransVOD++ and TransVOD Lite. For TransVOD, we adopt ResNet-50 as the backbone. For TransVOD++ and TransVOD Lite, we adopt Swin Transformer as the backbone.

\subsubsection{Ablation for TransVOD}

\noindent 
\textbf{Effectiveness of each component in TransVOD.}
Table~\ref{tab:transVOD_ablation_1}(a) summarizes the effects of different design components on the ImageNet VID dataset. 
% Temporal Query Encoder (TQE), Temporal Deformable Transformer Encoder (TDTE), and Temporal Deformable Transformer Decoder (TDTD) are three key components of our TransVOD. 
The single-frame baseline of Deformable DETR ~\cite{zhu2020deformable} is 76.0$\%$ and 88.3$\%$ with ResNet50 and Swin-Base Transformer, respectively. By merely using TDTE and TDTD, we boost the baseline with an additional +1.1$\%$ and +0.5$\%$ on the two backbones, respectively. Besides, by only adding TQE, we boost the baseline with an additional +2.9 $\%$, +1.0 $\%$ on the two backbones, respectively. The combination of TQE and TDTD increase the mAP to 79.3$\%$, 89.6$\%$, respectively. Finally, the proposed TransVOD including all components achieves 79.9$\%$ and 89.6$\%$ with ResNet50 and Swin-Base Transformer, respectively. These improvements show the effects of individual components of our TransVOD. Interestingly, we find the effect of TDTE fades away if we use a stronger backbone, \emph{e.g.,} Swin Transformer. 
%\vspace{2mm}

\noindent \textbf{Number of encoder layers in TDTE.}
Table~\ref{tab:transVOD_ablation_2}(a) illustrates the ablation study on the number of encoder layers in TDTE. We observe that when the number of TDTE encoder layers are larger than 1, it brings no significant benefits to the final performance. This experiment also proves the claim that aggregating the feature memories in a temporal dimension via deformable attention is useful for learning the temporal contexts across different frames.

\noindent 
\textbf{Number of encoder layers in TQE.}
Table~\ref{tab:transVOD_ablation_2}(b) shows the ablation study on the number of encoder layers in TQE. It shows that the best result occurs when the number query layer is set to 5. When the number of layers is up to 3, the performance is basically unchanged. Thus, we use 3 encoder layers in our final method.

%\vspace{2mm}
\noindent \textbf{Number of decoder layers in TDTD.}
Table~\ref{tab:transVOD_ablation_2}(c) illustrates the ablation study on the number of decoder layers in TDTD. The basic setting is 4 reference frames, 1 encoder layer in TQE, and 1 encoder layer in TDTE. The results indicate that only one decoder layer in TDTD is needed, and we set this number by default.
\begin{figure}[t]
    \centering
    \includegraphics[width=1.0\linewidth]{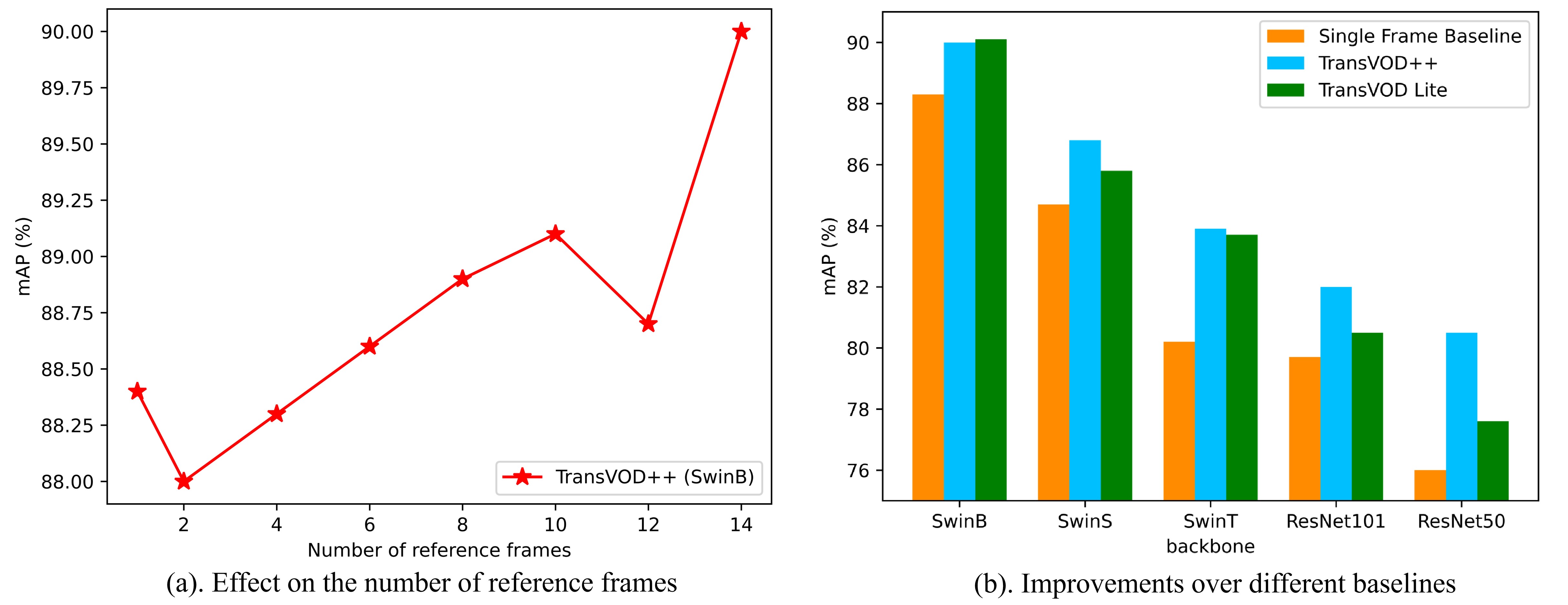} 
    \caption{\small Ablations of TransVOD++: (a). Effect on the number of reference frames $N_{ref}$ using Swin Base as the backbone. (b) Improvements over the different single frame baseline.}
    \label{fig:transvod_plus}
\end{figure}

\begin{table*}[h!]
    \footnotesize
 \centering
 \caption{Ablation studies of TransVOD++ on ImageNet VID using  Swin Transformer Base (SwinB) as the backbone.}
 \subfloat[Effect of each component of TransVOD++
    ]{
     \begin{tabularx}{0.35\textwidth}{l|c|c|c } 
        \toprule
        Component & (a) & (b) & (c) \\ \midrule
        Single Frame Baseline & \checkmark   & \checkmark & \checkmark\\
        RoI and Query Fusion &    & \checkmark & \checkmark\\ 
        Hard Query Mining &   &  & \checkmark \\  
        \midrule
        mAP$_{50}$ (\%) & 88.3  & 89.7   & \textbf{90.0} \\  \midrule
        mAP$_{50:95}$ (\%) &67.3 & 67.2  & \textbf{67.8} \\  \midrule
        mAP$_{50:95}$ (\%) (small) &14.2 & 16.0 & \textbf{17.6} \\  \midrule
        mAP$_{50:95}$ (\%) (medium) &39.0 & 41.5   & \textbf{42.1} \\  \midrule
        mAP$_{50:95}$ (\%) (large) &73.9 & 73.7 & \textbf{74.4} \\  \bottomrule
     \end{tabularx}
    }
\hfill
\subfloat[Effect of multi-level feature fusion.
    ]{
        %\label{tab:effect_component}
   \begin{tabularx}{0.3\textwidth}{l|c|c } 
        \toprule
        Component & (a) & (b)  \\ \midrule
        Single Frame Baseline & \checkmark  & \checkmark \\
        Multi-level feature fusion &    & \checkmark \\ 
        \midrule
        mAP$_{50}$ (\%) & 87.7   & 88.3 \\  \midrule
        mAP$_{50:95}$ (\%) & 65.0  &  67.3\\  \midrule
        mAP$_{50:95}$ (\%) (small)    & 12.2  & 14.2  \\  \midrule
        mAP$_{50:95}$ (\%) (medium)  & 35.3  & 39.0   \\  \midrule
        mAP$_{50:95}$ (\%) (large)    & 72.2  & 73.9 \\  \bottomrule
     \end{tabularx}
    }
\hfill
\subfloat[Effect of COCO pre-training.
    ]{
       \begin{tabularx}{0.3\textwidth}{l|c|c } 
        \toprule
        Component & (a) & (b)  \\ \midrule
        Single Frame Baseline & \checkmark  & \checkmark \\
        COCO pre-training &    & \checkmark \\ 
        \midrule
        mAP$_{50}$ (\%) & 44.8   & 88.3 \\  \midrule
        mAP$_{50:95}$ (\%) & 28.5  &  67.3\\  \midrule
        mAP$_{50:95}$ (\%) (small)    & 4.5  & 14.2  \\  \midrule
        mAP$_{50:95}$ (\%) (medium)  & 12.8  & 39.0   \\  \midrule
        mAP$_{50:95}$ (\%) (large)    & 33.3  & 73.9 \\  \bottomrule
     \end{tabularx}
     }
    \hfill
    \label{tab:ablation_using_transvod++}
    \vspace{-5mm}
\end{table*}

\begin{table*}[h!]
    \footnotesize
\centering
 \caption{Ablation studies of TransVOD Lite on ImageNet VID.}
 \hfill
  \subfloat[Effect of window size $T_w$ with Swin Tiny as backbone.
    ]{
     \begin{tabularx}{0.47\textwidth}{c|ccccccccc}
        \Xhline{1.0pt}
        $T_w$ & mAP$_{50}$ & mAP$_{50:95}$ & mAP$_S$ & mAP$_M$ & mAP$_L$ & FPS\\
        \hline
        1 & 76.6 & 55.1 &12.6 &31.5 &63.7 &16.5 \\
        2 & 79.1 & 56.7 &12.4 &34.1 &65.0 &21.7 \\
        4 & 80.9 & 57.9 &12.1 &35.1 &66.0 &23.5 \\
        6 & 81.5 & 58.3 &14.3 &35.8 &66.4 &22.9 \\
        8 & 82.1 & 58.6 &13.7 &36.3 &66.6 &22.5 \\
        10 & 82.3 & 58.7 &13.7 &35.9 &66.7 &29.2 \\
        12 & 82.7 & 59.0 &13.7 &\textbf{36.6} &\textbf{67.0} &30.1 \\
        14 & 82.5 & 58.8 &14.4 &\textbf{36.6} &66.8 &\textbf{32.2} \\
        15 & \textbf{83.7} & \textbf{66.2} &\textbf{14.7} &35.1 &67.3 &29.6 \\
        \Xhline{1.0pt}
     \end{tabularx}
    }
    \hfill
    \subfloat[Effect of window size $T_w$ with Swin Base as backbone.
    ]{
     \begin{tabularx}{0.47\textwidth}{c|ccccccccc}
        \Xhline{1.0pt}
        $T_w$ & mAP$_{50}$ & mAP$_{50:95}$ & mAP$_S$ & mAP$_M$ & mAP$_L$ & FPS\\
        \hline
        1 & 85.4 & 64.1 &13.8 &39.1 &72.0 &10.6 \\
        2 & 87.6 & 66.2 &14.2 &41.2 &74.0 &12.7 \\
        4 & 88.6 & 66.5 &\textbf{14.9} &42.4 &74.1 &14.2 \\
        6 & 89.2 & 67.0 &14.2 &42.3 &74.5 &15.2 \\
        8 & 88.9 & 66.7 &14.3 &42.6 &74.2 &15.4 \\
        10 & 88.8 & 66.4 &14.4 &42.6 &74.0 &15.6 \\
        12 & \textbf{90.1} & \textbf{67.7} &13.7 &\textbf{43.1} &\textbf{75.3} &\textbf{16.2} \\
        14 & 88.9 & 66.7 &14.4 &42.3 &74.2 &15.4 \\
        15 & 90.0 & 67.3 &\textbf{14.9} &41.6 &74.9 &15.0 \\
        \Xhline{1.0pt}
     \end{tabularx}
    }
    \hfill
    \subfloat[Effect of interval size $I_w$ using Swin Base as backbone.
    ]{
     \begin{tabularx}{0.45\textwidth}{c|ccccc}
        \Xhline{1.0pt}
        \multicolumn{6}{c}{(a) {Interval size $I_w$ when window size $T_w = 4$ .}}\\
        \hline
        $I_w (T_w = 4)$ & 1 & 4 & 8 & 12 & Randomly Shuffle\\
        mAP(\%) & 86.3 & 86.9 & 87.2 & 87.5 & 88.6 \\
        \Xhline{1.0pt}
        \multicolumn{6}{c}{(a) {Interval size $I_w$ when window size $T_w = 8$ .}}\\
        \hline
        $I_w (T_w  = 8)$ & 1 & 4 & 8 & 12 & Randomly Shuffle\\
        mAP(\%) & 86.6 & 87.3 & 88.0 &88.2  & 88.9 \\
        \Xhline{1.0pt}
        \multicolumn{6}{c}{(b) {Interval size $I_w$ when window size $T_w = 12$ .}}\\
        \hline
        $I_w (T_w  = 12)$ & 1 & 4 & 8 & 12 & Randomly Shuffle\\
        mAP(\%) & 86.9 & 88.0 & 88.7 & 89.3 & 90.1 \\
        \Xhline{1.0pt}
     \end{tabularx}
    }
    \hfill
    \subfloat[Ablation of top k query numbers in SeqHQM using ResNet-101 as backbone where the window size is set to 14.
    ]{
     \begin{tabularx}{0.43\textwidth}{c|cccccc} 
     \toprule[0.15em]
    k1 & 30 & 30 & 50  &80 & 100 & 100 \\
    k2 & 20 & 20 & 30  &50 & 80 & 80  \\
    k3 & 10 & 20 & 25  &30 & 50 & 30 \\
    \midrule[0.15em]
    mAP$_{50}$(\%) &78.4 &78.4 &78.8 & 80.4  &80.3  &79.9\\
    mAP$_{50:95}$(\%) &56.2 &56.4 & 56.5 &58.3   &58.2 &58.0 \\
    mAP$_{S}$(\%) &11.1 &11.3 & 11.4  &10.1  &10.0 &10.0 \\
    mAP$_{M}$(\%) &30.8 &31.2 & 31.4  &29.1 &29.3 &28.6 \\
    mAP$_{L}$(\%) &65.0 &65.1 & 65.2  &65.4  &65.2 &65.1 \\
    \bottomrule[0.15em]
     \end{tabularx}} 
    \label{tab:ablation_using_transvod_plus}
    \vspace{-5mm}
\end{table*}

% \vspace{2mm}
\noindent \textbf{Number of top $k$ object queries in TQE.}
To verify the effectiveness of our coarse-to-fine Temporal Query Aggregation strategy, we conduct ablation experiments in Table~\ref{tab:transVOD_ablation_2}(d) and Table~\ref{tab:transVOD_ablation_1}(b) to study how they contribute to the final performance. All the experiments in each table are conducted under the same setting. The first experiment is that when we use 1 encoder layer in TQE with 4 reference frames, the best performance is achieved when we choose the top 100 spatial object queries for each reference frame. The second experiment is conducted in a multiple TQE encoder layers case, \emph{i.e.,} 3 encoder layers in TQE. We denote the fine-to-fine (F2F) selection by using a small number of spatial object queries in each TQE encoder layer. Coarse-to-coarse (C2C)  means selecting a large number of spatial object queries when performing the aggregation in each layer. Our proposed coarse-to-fine (C2F) aggregation strategy uses larger number of spatial object queries in the shallow layers and a smaller number of spatial object queries in the deep layers. The results in Table~\ref{tab:transVOD_ablation_1}(b) show that our C2F aggregation strategy is superior to both the C2C and F2F selection. 

%\vspace{2mm}
\noindent \textbf{Number of reference frames in TransVOD.} Table~\ref{tab:transVOD_ablation_2}(d) illustrates the ablations on number of reference. The basic setting is 3 encoder layers in TQE, 1 encoder layer in TDTE, and 1 decoder layer in TDTD. As shown in Table~~\ref{tab:transVOD_ablation_2}(d), the mAP improves when the number of reference frames increases, and it tends to stabilize when the number is up to 8. Thus, in all experiments, we set the reference frames to 14 for TransVOD with different backbones.

\subsubsection{Ablation for TransVOD++}

\noindent 
\textbf{Effect of each component in TransVOD++ on strong baseline.} In Table~\ref{tab:ablation_using_transvod++}(a), we verify the effectiveness of each component in TransVOD++ on a strong baseline.
Adding RoI and Query Fusion results in 1.4$\%$ mAP improvements, while applying Hard Query Mining leads to extra 0.3$\%$ mAP improvements and 1.6$\%$ mAP improvements on small objects. This proves that our proposed Hard Query Mining is suitable for detecting small objects.

%\vspace{2mm}
\noindent
\textbf{Effect of reference frames in TransVOD++.} In Fig.~\ref{fig:transvod_plus}~(a), we show the effect of reference frames in TransVOD++ where we find the best reference frames is 14. This is different from the original TransVOD. We argue that utilizing more RoI information rather than full-frame fusion in the temporal dimension leads to better results. This finding is consistent with previous works~\cite{deng19rdn,han2021cfanet,chen2020memory} focusing on RoI-wised fusion in Faster-RCNN framework. We set the number of reference frames to 14 by default.

%\vspace{2mm}
\noindent
\textbf{Improvements over different baselines.}
In Fig.~\ref{fig:transvod_plus}~(b), we show the improvements over different single-frame baselines including Swin Transformer~\cite{liu2021swin} and ResNet~\cite{he16res}. Swin Base, Swin Small, and Swin Tiny are abbreviated as SwinB, SwinS, SwinT, respectively.
Our proposed TransVOD++ can boost the gain over 1.7\%$\sim$4.2\% mAP on various baselines. We observe that our TransVOD++ outperforms TransVOD Lite under different backbones. Especially, with ResNet-50 and ResNet-101, the improvements of TransVOD++ (2.3\%$\sim$4.2\% mAP ) are larger than the ones (0.8\%$\sim$1.6\%) mAP  of TransVOD Lite. Interestingly, we found that with the backbones of Swin Transformers, TransVOD Lite achieves almost the same performance as TransVOD++. This is mainly because the single frame baselines of Swin Transformer are too strong (88.3\%) mAP) and these improvements over the strong baseline are not as obvious as the ones of ResNet.

%\vspace{2mm}
\noindent
\textbf{Effect of multi-level feature fusion.} Table~\ref{tab:ablation_using_transvod++}(b) shows the improvements on multi-level feature fusion. In total, there is a 0.6$\%$ mAP$_{50}$ gain. Moreover, there is a more significant gain (2.3\%) on mAP$_{50:95}$ which indicates multi-scale information leads to more accurate detection results. Thus, we adopt the simple multi-level feature fusion by default when adopting Swin Transformer as the backbone for both TransVOD++ and TransVOD Lite.

%\vspace{2mm}
\noindent
\textbf{Effect of COCO pre-training using Swin base.} Following previous VOD methods~\cite{yao2020video,liu2019looking,qian2020adaptive}, we pre-train our image detector on the COCO dataset~\cite{COCO_dataset}. As shown in Table~\ref{tab:ablation_using_transvod++}(c), removing COCO pre-training leads to a huge performance drop. The main reason lies in the fact that vision Transformers~\cite{dosovitskiy2020image} need more training samples for better convergence and vision Transformers are typically pre-trained on large-scale datasets. 
Thus, we pre-train the TransVOD series on the COCO dataset. 

\subsubsection{Ablation for TransVOD Lite}
\label{sec:ablation_transvod_lite}

\begin{figure}[t!]
    \includegraphics[width=0.47\textwidth]{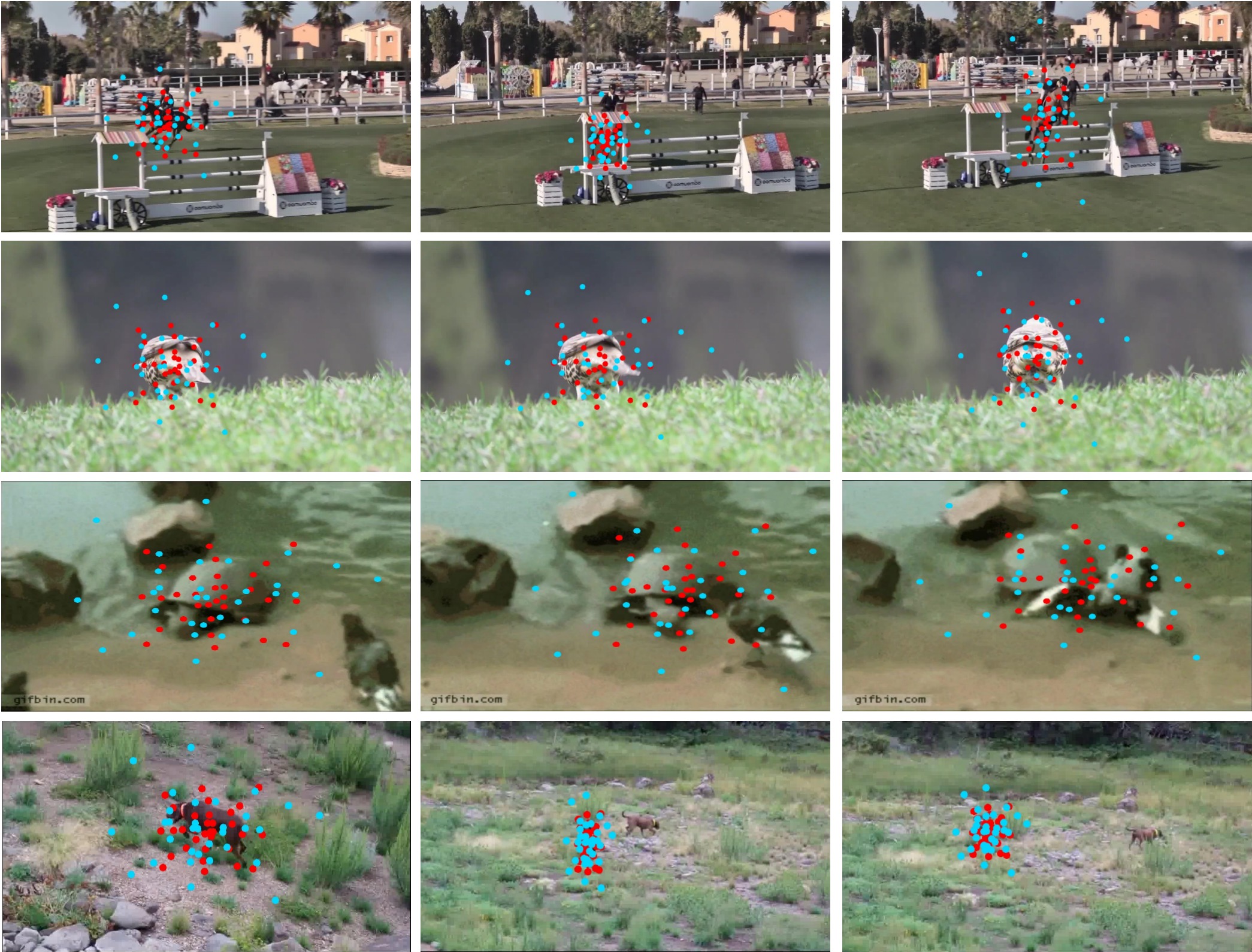} 
    \caption{\small The visualization of the deformable cross-attention in the last spatial Transformer decoder layer and temporal Transformer decoder layer. We visualize the sampling locations of the temporal object query and corresponding spatial object query in one figure. Each sampling point of the temporal object query is marked as a red-filled circle, while the blue circle represents the sampling point of the spatial query.}
    \label{fig:sample_vis_res}
\end{figure}

\begin{figure}[t!]
    \centering
    \includegraphics[width=1.0\linewidth]{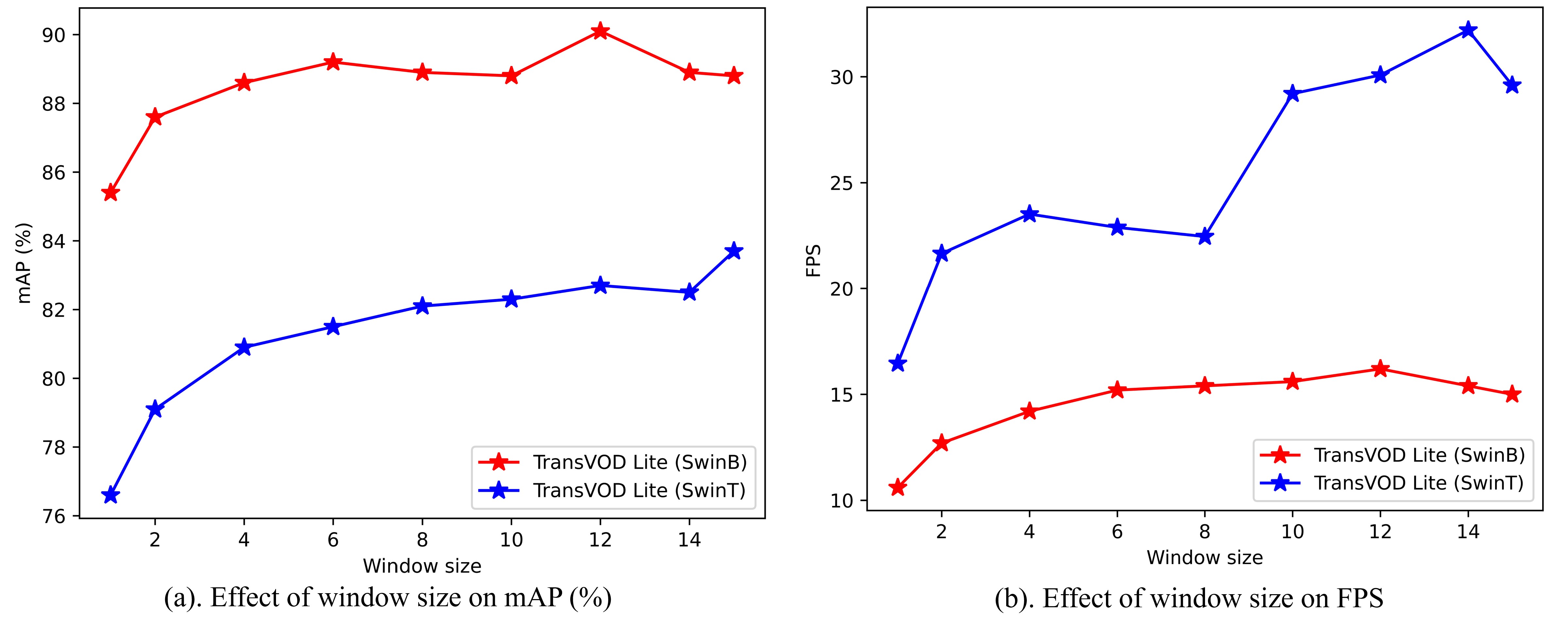} 
    \caption{\small Effect of the temporal window size $T_w$ of a video clip on the mean Average Precision (mAP) (a) and on the Frame Per Second (FPS) (b) in TransVOD Lite using Swin Base and Swin Tiny as the backbone, respectively. }
    \label{fig:transvod_lite}
    \vspace{-5mm}
\end{figure}

\noindent
\textbf{Effect of window size in TransVOD Lite.} In Fig.~\ref{fig:transvod_lite}~(a) and Fig.~\ref{fig:transvod_lite}~(b), we show the effect of window size on both accuracy and inference time where the interval mode is randomly shuffled within the window for all experiments. As shown in these figures, increasing window size leads to both accuracy improvements and FPS increase for both Swin Tiny and Swin base as backbones. In Table~\ref{tab:ablation_using_transvod_plus}(a) and Table~\ref{tab:ablation_using_transvod_plus}(b), we detail the results of the above figures. We choose the best window size $T_{w}$ as 15 for all models.

%\vspace{2mm}
\noindent
\textbf{Effect of interval size and mode in TransVOD Lite.} 
In Table~\ref{tab:ablation_using_transvod_plus}(c), we show the effect of interval size between frames in each fixed window. For different window sizes, increasing the interval size leads to better results. This indicates that fusing more global temporal information leads to better results. However, adopting our proposed randomly shuffled strategy results in the best performance on different window sizes. This is mainly because random shuffles increase the diversity of each frame. For example, the global and local temporal information can exist in one window. Moreover, during training, the frames are randomly selected from each clip. Thus, randomly shuffled inputs share the same distribution with training examples. We report the final performance using such settings. Moreover, as shown in Table~\ref{tab:ablation_using_transvod_plus}(c), even with the sequential inputs, our methods can still achieve the best performance compared with the methods in Table~\ref{table:transvod_lite_stoa}.

%\vspace{2mm}
\noindent
\textbf{Ablation on query numbers in Sequential Hard Query Mining.} In Table~\ref{tab:ablation_using_transvod_plus}(d), we perform ablation studies on Sequential Hard Query Mining (SeqHQM) in TransVOD Lite. From the table, we find the best hyper-parameter with 80, 50, 30 queries for each stage. We use that setting for all the TransVOD Lite models.

\begin{figure}[t]
    \centering
    \includegraphics[width=0.95\linewidth]{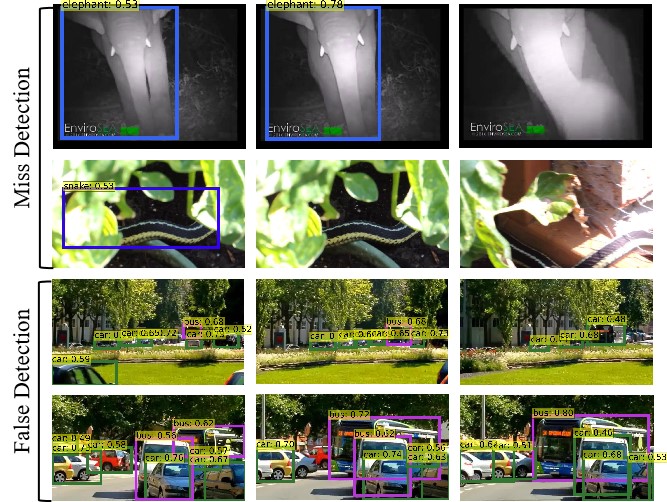} 
    \caption{\small Failure case analysis. First and second row: miss detection. Third and fourth row: false detection. The results are obtained via our TransVOD Lite with Swin Base backbone.
    }
    \label{fig:failure_cases}
\end{figure}

\begin{figure*}[h]
    \centering
    \includegraphics[width=0.95\textwidth]{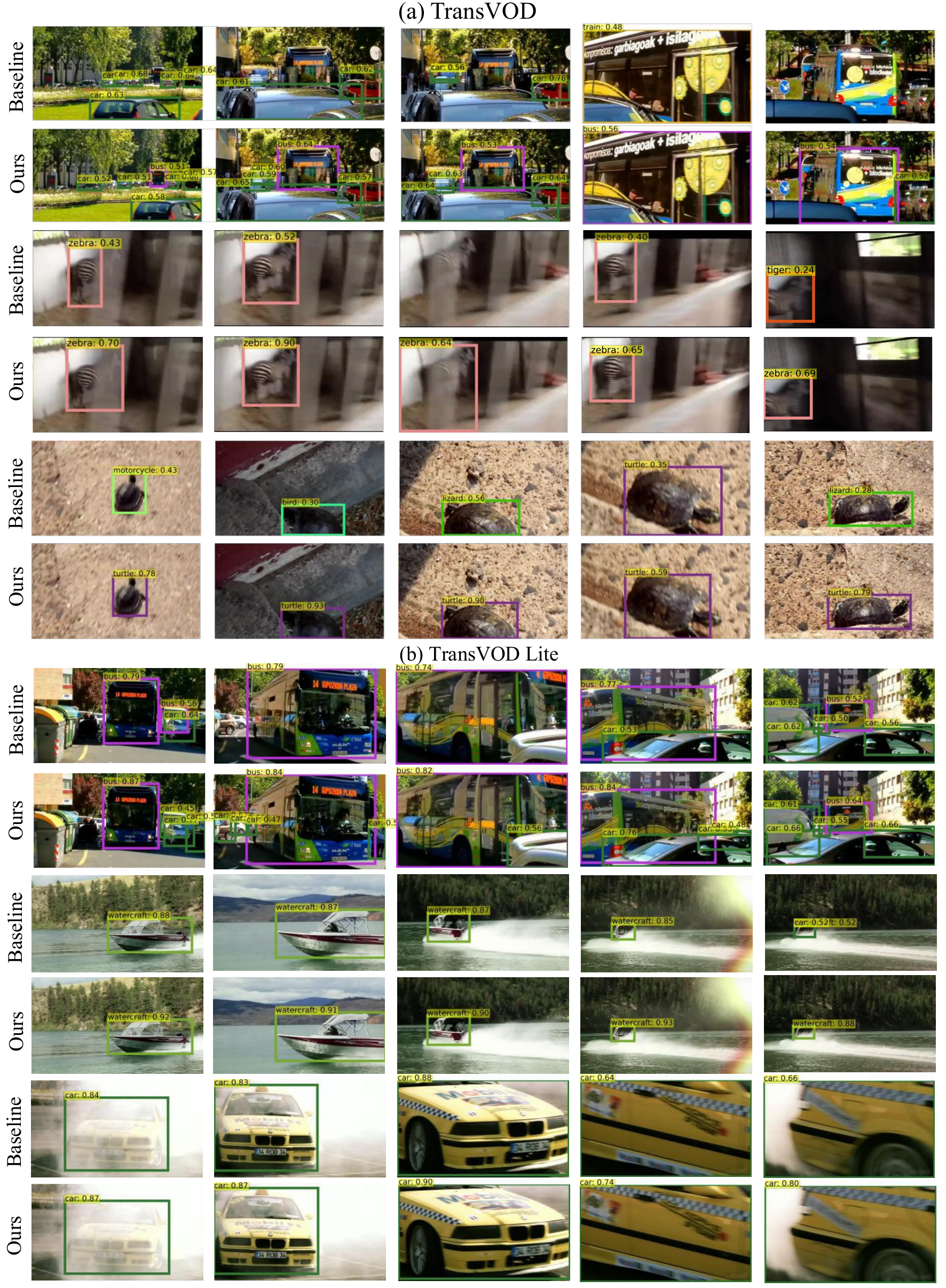} 
    \vspace{-3mm}
    \caption{\small The visualization results of single frame baseline method~\cite{zhu2020deformable} and TransVOD (a), TransVOD Lite (b) in different scenarios. Compared with single frame baseline, our proposed TransVOD and TransVOD lite show better and consistent detection results in the cases of part occlusion (top two rows of (a) and (b)), motion blur (middle two rows of (a) and (b)) and rare pose (last two rows of (a) and (b)), respectively. Best view it on the screen and zoom in.}
    \label{fig:visual_prediction_results}
    \vspace{-5mm}
\end{figure*}

\subsection{Visualization and Analysis}

\noindent \textbf{Visual detection results.} As shown in Fig.~\ref{fig:visual_prediction_results}, we show the visual detection results of still image detector, \emph{i.e.,} Deformable DETR~\cite{zhu2020deformable} and our proposed TransVOD in odd and even rows, respectively. The still image detector is easy to cause false detection (\emph{e.g.,} turtle detected as a lizard) and missed detection (\emph{e.g.,} zebra not detected), in the case of motion blur, part occlusion. Compared with Deformable DETR~\cite{zhu2020deformable}, our method effectively models the long-range dependencies across different video frames to enhance the features of the detected image. Thus, our TransVOD not only increases the confidence of correct prediction, but also effectively reduces the number of cases that are missed or falsely detected. Besides, as shown in Fig.~\ref{fig:visual_prediction_results} (b), our TransVOD Lite shows more confident scores than the single frame baseline \cite{zhu2020deformable}.

\noindent \textbf{Visual sampling locations of object query in TransVOD.} To further explore the advantages of TQE, we visualize the sampling locations of both spatial object query and temporal object query in Fig.~\ref{fig:sample_vis_res}. The sample locations indicate the most relevant context for each detection. As shown in the figure, for each frame in each clip, our temporal object query \textit{has more concentrated and precise results} on foreground objects while the original spatial object query has more diffuse results. This proves that our temporal object query is more suitable for detecting objects in video. This explains the effectiveness of our temporal query fusion.

\noindent \textbf{Failure case analysis.}
In Fig.~\ref{fig:failure_cases}, we present several failure cases using our best TransVOD Lite model. The first two rows show the missing detection problems. The first row is mainly due to the larger motion blur, and the second row is caused by the various background change. The last two rows show the false detection where a car is detected as a bus. This is caused by the large occlusion. Both cases show that tackling occlusion and more stable temporal modeling are needed for the further works.

\section{Conclusion}
In this paper, we proposed a novel video object detection framework, namely TransVOD, which provides a new perspective of feature aggregation by leveraging spatial-temporal Transformers. TransVOD effectively removes the need for many hand-crafted components and complicated post-processing methods. Our core idea is to aggregate both the spatial object queries and the memory encodings in each frame via temporal Transformers. Our TransVOD boosts the strong baseline deformable DETR by a significant margin (3\%-4\% mAP) on the ImageNet VID dataset on various baselines. To our knowledge, our work is the \textbf{first one} that applies the Transformers to VOD. Based on the TransVOD framework, we present two advanced versions, namely TransVOD++ and TransVOD Lite. The former improves the performance of TransVOD via better Query and RoI fusion (QRF), and Hard Query Mining (HQM) to fully utilize the object-level information, and dynamically reduce the number of object queries and targets. The latter focuses on real-time video object detection by modeling VOD as a sequence-to-sequence prediction problem via Sequential Hard Query Mining (SeqHQM). Both models set new state-of-the-art results on the ImageNet VID dataset on two different settings: accuracy for non-real-time models and best speed-accuracy trade-off on real-time models. Our method is the first work that achieves 90$\%$ mAP on ImageNet VID dataset. Moreover, TransVOD is the first work that achieves 83.7\% mAP while running in real time. We believe our models can be new baselines for this area.

% \section*{Acknowledgment}
% This work is supported by National Key Research and Development Program of China (2019YFC1521104), National Natural Science Foundation of China (72192821, 61972157), Shanghai Municipal Science and Technology Major Project  (2021SHZDZX0102), Shanghai Science and Technology Commission (21511101200, 22YF1420300), and Art major project of National Social Science Fund (I8ZD22). Lizhuang Ma is corresponding author.

\ifCLASSOPTIONcaptionsoff
  \newpage
\fi

\bibliographystyle{IEEEtran}
\bibliography{main}

\begin{thebibliography}{10}
\providecommand{\url}[1]{#1}
\csname url@samestyle\endcsname
\providecommand{\newblock}{\relax}
\providecommand{\bibinfo}[2]{#2}
\providecommand{\BIBentrySTDinterwordspacing}{\spaceskip=0pt\relax}
\providecommand{\BIBentryALTinterwordstretchfactor}{4}
\providecommand{\BIBentryALTinterwordspacing}{\spaceskip=\fontdimen2\font plus
\BIBentryALTinterwordstretchfactor\fontdimen3\font minus
  \fontdimen4\font\relax}
\providecommand{\BIBforeignlanguage}[2]{{%
\expandafter\ifx\csname l@#1\endcsname\relax
\typeout{** WARNING: IEEEtran.bst: No hyphenation pattern has been}%
\typeout{** loaded for the language `#1'. Using the pattern for}%
\typeout{** the default language instead.}%
\else
\language=\csname l@#1\endcsname
\fi
#2}}
\providecommand{\BIBdecl}{\relax}
\BIBdecl

\bibitem{ren2016faster}
S.~Ren, K.~He, R.~Girshick, and J.~Sun, ``Faster r-cnn: towards real-time
  object detection with region proposal networks,'' \emph{IEEE transactions on
  pattern analysis and machine intelligence}, vol.~39, no.~6, pp. 1137--1149,
  2016.

\bibitem{dai16rfcn}
J.~Dai, Y.~Li, K.~He, and J.~Sun, ``R-fcn: Object detection via region-based
  fully convolutional networks,'' in \emph{Advances in neural information
  processing systems}, 2016, pp. 379--387.

\bibitem{FocalLoss}
T.-Y. Lin, P.~Goyal, R.~Girshick, K.~He, and P.~Doll{\'a}r, ``Focal loss for
  dense object detection,'' in \emph{Proceedings of the IEEE international
  conference on computer vision}, 2017, pp. 2980--2988.

\bibitem{tian2019fcos}
Z.~Tian, C.~Shen, H.~Chen, and T.~He, ``Fcos: Fully convolutional one-stage
  object detection,'' in \emph{Proceedings of the IEEE/CVF International
  Conference on Computer Vision}, 2019, pp. 9627--9636.

\bibitem{han2016seq}
W.~Han, P.~Khorrami, T.~L. Paine, P.~Ramachandran, M.~Babaeizadeh, H.~Shi,
  J.~Li, S.~Yan, and T.~S. Huang, ``Seq-nms for video object detection,''
  \emph{arXiv preprint arXiv:1602.08465}, 2016.

\bibitem{kang2017t}
K.~Kang, H.~Li, J.~Yan, X.~Zeng, B.~Yang, T.~Xiao, C.~Zhang, Z.~Wang, R.~Wang,
  X.~Wang \emph{et~al.}, ``T-cnn: Tubelets with convolutional neural networks
  for object detection from videos,'' \emph{IEEE Transactions on Circuits and
  Systems for Video Technology}, vol.~28, no.~10, pp. 2896--2907, 2017.

\bibitem{belhassen2019improving}
H.~Belhassen, H.~Zhang, V.~Fresse, and E.-B. Bourennane, ``Improving video
  object detection by seq-bbox matching.'' in \emph{VISIGRAPP (5: VISAPP)},
  2019, pp. 226--233.

\bibitem{sabater2020robust}
A.~Sabater, L.~Montesano, and A.~C. Murillo, ``Robust and efficient
  post-processing for video object detection,'' \emph{arXiv preprint
  arXiv:2009.11050}, 2020.

\bibitem{yao2020video}
C.-H. Yao, C.~Fang, X.~Shen, Y.~Wan, and M.-H. Yang, ``Video object detection
  via object-level temporal aggregation,'' in \emph{European conference on
  computer vision}.\hskip 1em plus 0.5em minus 0.4em\relax Springer, 2020, pp.
  160--177.

\bibitem{jiang2020learning}
Z.~Jiang, Y.~Liu, C.~Yang, J.~Liu, P.~Gao, Q.~Zhang, S.~Xiang, and C.~Pan,
  ``Learning where to focus for efficient video object detection,'' in
  \emph{European Conference on Computer Vision}.\hskip 1em plus 0.5em minus
  0.4em\relax Springer, 2020, pp. 18--34.

\bibitem{han2020mining}
M.~Han, Y.~Wang, X.~Chang, and Y.~Qiao, ``Mining inter-video proposal relations
  for video object detection,'' in \emph{European Conference on Computer
  Vision}.\hskip 1em plus 0.5em minus 0.4em\relax Springer, 2020, pp. 431--446.

\bibitem{han2020exploiting}
L.~Han, P.~Wang, Z.~Yin, F.~Wang, and H.~Li, ``Exploiting better feature
  aggregation for video object detection,'' in \emph{Proceedings of the 28th
  ACM International Conference on Multimedia}, 2020, pp. 1469--1477.

\bibitem{lin2020dual}
L.~Lin, H.~Chen, H.~Zhang, J.~Liang, Y.~Li, Y.~Shan, and H.~Wang, ``Dual
  semantic fusion network for video object detection,'' in \emph{Proceedings of
  the 28th ACM International Conference on Multimedia}, 2020, pp. 1855--1863.

\bibitem{he2020temporal}
F.~He, N.~Gao, Q.~Li, S.~Du, X.~Zhao, and K.~Huang, ``Temporal context enhanced
  feature aggregation for video object detection,'' in \emph{Proceedings of the
  AAAI Conference on Artificial Intelligence}, vol.~34, no.~07, 2020, pp.
  10\,941--10\,948.

\bibitem{chen2018optimizing}
K.~Chen, J.~Wang, S.~Yang, X.~Zhang, Y.~Xiong, C.~C. Loy, and D.~Lin,
  ``Optimizing video object detection via a scale-time lattice,'' in
  \emph{Proceedings of the IEEE conference on computer vision and pattern
  recognition}, 2018, pp. 7814--7823.

\bibitem{chen2020memory}
Y.~Chen, Y.~Cao, H.~Hu, and L.~Wang, ``Memory enhanced global-local aggregation
  for video object detection,'' in \emph{Proceedings of the IEEE/CVF Conference
  on Computer Vision and Pattern Recognition}, 2020, pp. 10\,337--10\,346.

\bibitem{sun2021mamba}
G.~Sun, Y.~Hua, G.~Hu, and N.~Robertson, ``Mamba: Multi-level aggregation via
  memory bank for video object detection,'' in \emph{Proceedings of the AAAI
  Conference on Artificial Intelligence}, vol.~35, no.~3, 2021, pp. 2620--2627.

\bibitem{guo2019progressive}
C.~Guo, B.~Fan, J.~Gu, Q.~Zhang, S.~Xiang, V.~Prinet, and C.~Pan, ``Progressive
  sparse local attention for video object detection,'' in \emph{Proceedings of
  the IEEE/CVF International Conference on Computer Vision}, 2019, pp.
  3909--3918.

\bibitem{zhu17dff}
X.~Zhu, Y.~Xiong, J.~Dai, L.~Yuan, and Y.~Wei, ``Deep feature flow for video
  recognition,'' in \emph{Proceedings of the IEEE conference on computer vision
  and pattern recognition}, 2017, pp. 2349--2358.

\bibitem{zhu17fgfa}
X.~Zhu, Y.~Wang, J.~Dai, L.~Yuan, and Y.~Wei, ``Flow-guided feature aggregation
  for video object detection,'' in \emph{Proceedings of the IEEE International
  Conference on Computer Vision}, 2017, pp. 408--417.

\bibitem{wang18manet}
S.~Wang, Y.~Zhou, J.~Yan, and Z.~Deng, ``Fully motion-aware network for video
  object detection,'' in \emph{Proceedings of the European conference on
  computer vision (ECCV)}, 2018, pp. 542--557.

\bibitem{zhu18hp}
X.~Zhu, J.~Dai, L.~Yuan, and Y.~Wei, ``Towards high performance video object
  detection,'' in \emph{Proceedings of the IEEE Conference on Computer Vision
  and Pattern Recognition}, 2018, pp. 7210--7218.

\bibitem{jin2022feature}
R.~Jin, G.~Lin, C.~Wen, J.~Wang, and F.~Liu, ``Feature flow: In-network feature
  flow estimation for video object detection,'' \emph{Pattern Recognition},
  vol. 122, p. 108323, 2022.

\bibitem{deng2019ogemn}
H.~Deng, Y.~Hua, T.~Song, Z.~Zhang, Z.~Xue, R.~Ma, N.~Robertson, and H.~Guan,
  ``Object guided external memory network for video object detection,'' in
  \emph{Proceedings of the IEEE/CVF International Conference on Computer
  Vision}, 2019, pp. 6678--6687.

\bibitem{bertasius18stsn}
G.~Bertasius, L.~Torresani, and J.~Shi, ``Object detection in video with
  spatiotemporal sampling networks,'' in \emph{Proceedings of the European
  Conference on Computer Vision (ECCV)}, 2018, pp. 331--346.

\bibitem{jiang2019video}
Z.~Jiang, P.~Gao, C.~Guo, Q.~Zhang, S.~Xiang, and C.~Pan, ``Video object
  detection with locally-weighted deformable neighbors,'' in \emph{Proceedings
  of the AAAI Conference on Artificial Intelligence}, vol.~33, no.~01, 2019,
  pp. 8529--8536.

\bibitem{deng19rdn}
J.~Deng, Y.~Pan, T.~Yao, W.~Zhou, H.~Li, and T.~Mei, ``Relation distillation
  networks for video object detection,'' in \emph{Proceedings of the IEEE/CVF
  International Conference on Computer Vision}, 2019, pp. 7023--7032.

\bibitem{shvets19lltr}
M.~Shvets, W.~Liu, and A.~C. Berg, ``Leveraging long-range temporal
  relationships between proposals for video object detection,'' in
  \emph{Proceedings of the IEEE/CVF International Conference on Computer
  Vision}, 2019, pp. 9756--9764.

\bibitem{liu2019looking}
M.~Liu, M.~Zhu, M.~White, Y.~Li, and D.~Kalenichenko, ``Looking fast and slow:
  Memory-guided mobile video object detection,'' \emph{arXiv preprint
  arXiv:1903.10172}, 2019.

\bibitem{liu2018mobile}
M.~Liu and M.~Zhu, ``Mobile video object detection with temporally-aware
  feature maps,'' in \emph{Proceedings of the IEEE conference on computer
  vision and pattern recognition}, 2018, pp. 5686--5695.

\bibitem{Chen2018OptimizingVOD}
K.~Chen, J.~Wang, S.~Yang, X.~Zhang, Y.~Xiong, C.~C. Loy, and D.~Lin,
  ``Optimizing video object detection via a scale-time lattice,'' \emph{2018
  IEEE/CVF Conference on Computer Vision and Pattern Recognition}, pp.
  7814--7823, 2018.

\bibitem{wang2020end}
Y.~Wang, Z.~Xu, X.~Wang, C.~Shen, B.~Cheng, H.~Shen, and H.~Xia, ``End-to-end
  video instance segmentation with transformers,'' \emph{arXiv preprint
  arXiv:2011.14503}, 2020.

\bibitem{detr}
N.~Carion, F.~Massa, G.~Synnaeve, N.~Usunier, A.~Kirillov, and S.~Zagoruyko,
  ``End-to-end object detection with transformers,'' in \emph{European
  Conference on Computer Vision}.\hskip 1em plus 0.5em minus 0.4em\relax
  Springer, 2020, pp. 213--229.

\bibitem{zhu2020deformable}
X.~Zhu, W.~Su, L.~Lu, B.~Li, X.~Wang, and J.~Dai, ``Deformable detr: Deformable
  transformers for end-to-end object detection,'' \emph{arXiv preprint
  arXiv:2010.04159}, 2020.

\bibitem{dosovitskiy2020image}
A.~Dosovitskiy, L.~Beyer, A.~Kolesnikov, D.~Weissenborn, X.~Zhai,
  T.~Unterthiner, M.~Dehghani, M.~Minderer, G.~Heigold, S.~Gelly \emph{et~al.},
  ``An image is worth 16x16 words: Transformers for image recognition at
  scale,'' \emph{The International Conference on Learning Representations
  (ICLR)}, 2021.

\bibitem{sun2020transtrack}
P.~Sun, Y.~Jiang, R.~Zhang, E.~Xie, J.~Cao, X.~Hu, T.~Kong, Z.~Yuan, C.~Wang,
  and P.~Luo, ``Transtrack: Multiple-object tracking with transformer,''
  \emph{arXiv preprint arXiv:2012.15460}, 2020.

\bibitem{Vaswani17attention}
A.~Vaswani, N.~Shazeer, N.~Parmar, J.~Uszkoreit, L.~Jones, A.~N. Gomez,
  {\L}.~Kaiser, and I.~Polosukhin, ``Attention is all you need,'' in
  \emph{Advances in neural information processing systems}, 2017, pp.
  5998--6008.

\bibitem{russakovsky2015imagenet}
O.~Russakovsky, J.~Deng, H.~Su, J.~Krause, S.~Satheesh, S.~Ma, Z.~Huang,
  A.~Karpathy, A.~Khosla, M.~Bernstein \emph{et~al.}, ``Imagenet large scale
  visual recognition challenge,'' \emph{International journal of computer
  vision}, vol. 115, no.~3, pp. 211--252, 2015.

\bibitem{he2021end}
L.~He, Q.~Zhou, X.~Li, L.~Niu, G.~Cheng, X.~Li, W.~Liu, Y.~Tong, L.~Ma, and
  L.~Zhang, ``End-to-end video object detection with spatial-temporal
  transformers,'' in \emph{Proceedings of the 29th ACM International Conference
  on Multimedia (ACM MM)}, 2021, p. 1507–1516.

\bibitem{ohem}
A.~Shrivastava, A.~Gupta, and R.~Girshick, ``Training region-based object
  detectors with online hard example mining,'' in \emph{Proceedings of the IEEE
  conference on computer vision and pattern recognition}, 2016, pp. 761--769.

\bibitem{SegOHEM}
Z.~Wu, C.~Shen, and A.~v.~d. Hengel, ``High-performance semantic segmentation
  using very deep fully convolutional networks,'' \emph{arXiv preprint}, 2016.

\bibitem{liu2021swin}
Z.~Liu, Y.~Lin, Y.~Cao, H.~Hu, Y.~Wei, Z.~Zhang, S.~Lin, and B.~Guo, ``Swin
  transformer: Hierarchical vision transformer using shifted windows,''
  \emph{ICCV}, 2021.

\bibitem{dosovitskiy2015flownet}
A.~Dosovitskiy, P.~Fischer, E.~Ilg, P.~Hausser, C.~Hazirbas, V.~Golkov, P.~Van
  Der~Smagt, D.~Cremers, and T.~Brox, ``Flownet: Learning optical flow with
  convolutional networks,'' in \emph{Proceedings of the IEEE international
  conference on computer vision}, 2015, pp. 2758--2766.

\bibitem{wu19selsa}
H.~Wu, Y.~Chen, N.~Wang, and Z.~Zhang, ``Sequence level semantics aggregation
  for video object detection,'' in \emph{Proceedings of the IEEE/CVF
  International Conference on Computer Vision}, 2019, pp. 9217--9225.

\bibitem{vaswani2017attention}
A.~Vaswani, N.~Shazeer, N.~Parmar, J.~Uszkoreit, L.~Jones, A.~N. Gomez,
  L.~Kaiser, and I.~Polosukhin, ``Attention is all you need,'' \emph{arXiv
  preprint arXiv:1706.03762}, 2017.

\bibitem{wang2018non}
X.~Wang, R.~Girshick, A.~Gupta, and K.~He, ``Non-local neural networks,'' in
  \emph{Proceedings of the IEEE conference on computer vision and pattern
  recognition}, 2018, pp. 7794--7803.

\bibitem{chin2019adascale}
T.-W. Chin, R.~Ding, and D.~Marculescu, ``Adascale: Towards real-time video
  object detection using adaptive scaling,'' \emph{arXiv preprint
  arXiv:1902.02910}, 2019.

\bibitem{li2021improving}
X.~Li, H.~He, H.~Ding, K.~Yang, G.~Cheng, J.~Shi, and Y.~Tong, ``Improving
  video instance segmentation via temporal pyramid routing,'' \emph{arXiv
  preprint arXiv:2107.13155}, 2021.

\bibitem{Chai2019PatchworkAP}
Y.~Chai, ``Patchwork: A patch-wise attention network for efficient object
  detection and segmentation in video streams,'' \emph{2019 IEEE/CVF
  International Conference on Computer Vision (ICCV)}, pp. 3414--3423, 2019.

\bibitem{meinhardt2021trackformer}
T.~Meinhardt, A.~Kirillov, L.~Leal-Taixe, and C.~Feichtenhofer, ``Trackformer:
  Multi-object tracking with transformers,'' \emph{arXiv preprint
  arXiv:2101.02702}, 2021.

\bibitem{deit_vit}
H.~Touvron, M.~Cord, M.~Douze, F.~Massa, A.~Sablayrolles, and H.~J{\'e}gou,
  ``Training data-efficient image transformers \& distillation through
  attention,'' in \emph{International Conference on Machine Learning}.\hskip
  1em plus 0.5em minus 0.4em\relax PMLR, 2021, pp. 10\,347--10\,357.

\bibitem{zhang2022eatformer}
J.~Zhang, X.~Li, Y.~Wang, C.~Wang, Y.~Yang, Y.~Liu, and D.~Tao, ``Eatformer:
  Improving vision transformer inspired by evolutionary algorithm,''
  \emph{arXiv preprint arXiv:2206.09325}, 2022.

\bibitem{video_knet}
X.~Li, W.~Zhang, J.~Pang, K.~Chen, G.~Cheng, Y.~Tong, and C.~C. Loy, ``Video
  k-net: A simple, strong, and unified baseline for video segmentation,'' in
  \emph{CVPR}, 2022.

\bibitem{panopticpartformer}
X.~Li, S.~Xu, Y.~Yang, G.~Cheng, Y.~Tong, and D.~Tao, ``Panoptic-partformer:
  Learning a unified model for panoptic part segmentation,'' in \emph{Eur.
  Conf. Comput. Vis.}, 2022.

\bibitem{fashionformer}
S.~Xu, X.~Li, J.~Wang, G.~Cheng, Y.~Tong, and D.~Tao, ``Fashionformer: A
  simple, effective and unified baseline for human fashion segmentation and
  recognition,'' in \emph{Eur. Conf. Comput. Vis.}, 2022.

\bibitem{vis_dataset}
L.~Yang, Y.~Fan, and N.~Xu, ``Video instance segmentation,'' in
  \emph{Proceedings of the IEEE/CVF International Conference on Computer
  Vision}, 2019, pp. 5188--5197.

\bibitem{he16res}
K.~He, X.~Zhang, S.~Ren, and J.~Sun, ``Deep residual learning for image
  recognition,'' in \emph{Proceedings of the IEEE conference on computer vision
  and pattern recognition}, 2016, pp. 770--778.

\bibitem{dai2017deformable}
J.~Dai, H.~Qi, Y.~Xiong, Y.~Li, G.~Zhang, H.~Hu, and Y.~Wei, ``Deformable
  convolutional networks,'' in \emph{Proceedings of the IEEE international
  conference on computer vision}, 2017, pp. 764--773.

\bibitem{hu18relationnet}
H.~Hu, J.~Gu, Z.~Zhang, J.~Dai, and Y.~Wei, ``Relation networks for object
  detection,'' in \emph{Proceedings of the IEEE conference on computer vision
  and pattern recognition}, 2018, pp. 3588--3597.

\bibitem{sun2021sparse}
P.~Sun, R.~Zhang, Y.~Jiang, T.~Kong, C.~Xu, W.~Zhan, M.~Tomizuka, L.~Li,
  Z.~Yuan, C.~Wang \emph{et~al.}, ``Sparse r-cnn: End-to-end object detection
  with learnable proposals,'' in \emph{Proceedings of the IEEE/CVF Conference
  on Computer Vision and Pattern Recognition}, 2021, pp. 14\,454--14\,463.

\bibitem{he17maskrcnn}
K.~He, G.~Gkioxari, P.~Doll{\'a}r, and R.~Girshick, ``Mask r-cnn,'' in
  \emph{Proceedings of the IEEE international conference on computer vision},
  2017, pp. 2961--2969.

\bibitem{stewart2016end}
R.~Stewart, M.~Andriluka, and A.~Y. Ng, ``End-to-end people detection in
  crowded scenes,'' in \emph{Proceedings of the IEEE conference on computer
  vision and pattern recognition}, 2016, pp. 2325--2333.

\bibitem{kuhn1955hungarian}
H.~W. Kuhn, ``The hungarian method for the assignment problem,'' \emph{Naval
  research logistics quarterly}, vol.~2, no. 1-2, pp. 83--97, 1955.

\bibitem{GIoU}
H.~Rezatofighi, N.~Tsoi, J.~Gwak, A.~Sadeghian, I.~Reid, and S.~Savarese,
  ``Generalized intersection over union: A metric and a loss for bounding box
  regression,'' in \emph{Proceedings of the IEEE/CVF Conference on Computer
  Vision and Pattern Recognition}, 2019, pp. 658--666.

\bibitem{qian2020adaptive}
Y.~Qian, L.~Yu, W.~Liu, G.~Kang, and A.~G. Hauptmann, ``Adaptive feature
  aggregation for video object detection,'' in \emph{Proceedings of the
  IEEE/CVF Winter Conference on Applications of Computer Vision Workshops},
  2020, pp. 143--147.

\bibitem{COCO_dataset}
T.-Y. Lin, M.~Maire, S.~Belongie, J.~Hays, P.~Perona, D.~Ramanan,
  P.~Doll{\'a}r, and C.~L. Zitnick, ``Microsoft coco: Common objects in
  context,'' in \emph{European Conference on Computer Vision}, 2014.

\bibitem{deng2009imagenet}
J.~Deng, W.~Dong, R.~Socher, L.-J. Li, K.~Li, and L.~Fei-Fei, ``Imagenet: A
  large-scale hierarchical image database,'' in \emph{2009 IEEE conference on
  computer vision and pattern recognition}.\hskip 1em plus 0.5em minus
  0.4em\relax IEEE, 2009, pp. 248--255.

\bibitem{loshchilov2017decoupled}
I.~Loshchilov and F.~Hutter, ``Decoupled weight decay regularization,''
  \emph{arXiv preprint arXiv:1711.05101}, 2017.

\bibitem{glorot2010understanding}
X.~Glorot and Y.~Bengio, ``Understanding the difficulty of training deep
  feedforward neural networks,'' in \emph{Proceedings of the thirteenth
  international conference on artificial intelligence and statistics}, 2010,
  pp. 249--256.

\bibitem{mmtrack2020}
M.~Contributors, ``{MMTracking: OpenMMLab} video perception toolbox and
  benchmark,'' \url{https://github.com/open-mmlab/mmtracking}, 2020.

\bibitem{feichtenhofer17dt}
C.~Feichtenhofer, A.~Pinz, and A.~Zisserman, ``Detect to track and track to
  detect,'' in \emph{Proceedings of the IEEE international conference on
  computer vision}, 2017, pp. 3057--3065.

\bibitem{dai2016r}
J.~Dai, Y.~Li, K.~He, and J.~Sun, ``R-fcn: Object detection via region-based
  fully convolutional networks,'' \emph{arXiv preprint arXiv:1605.06409}, 2016.

\bibitem{wang2020scnet}
F.~Wang, Z.~Xu, Y.~Gan, C.-M. Vong, and Q.~Liu, ``Scnet: Scale-aware
  coupling-structure network for efficient video object detection,''
  \emph{Neurocomputing}, vol. 404, pp. 283--293, 2020.

\bibitem{xiao18stmn}
F.~Xiao and Y.~J. Lee, ``Video object detection with an aligned
  spatial-temporal memory,'' in \emph{Proceedings of the European Conference on
  Computer Vision (ECCV)}, 2018, pp. 485--501.

\bibitem{wu2020bfan}
Y.~Wu, H.~Zhang, Y.~Li, Y.~Yang, and D.~Yuan, ``Video object detection guided
  by object blur evaluation,'' \emph{IEEE Access}, vol.~8, pp.
  208\,554--208\,565, 2020.

\bibitem{luo2019object}
H.~Luo, L.~Huang, H.~Shen, Y.~Li, C.~Huang, and X.~Wang, ``Object detection in
  video with spatial-temporal context aggregation,'' \emph{arXiv preprint
  arXiv:1907.04988}, 2019.

\bibitem{MINet}
J.~Deng, Y.~Pan, T.~Yao, W.~Zhou, H.~Li, and T.~Mei, ``Minet: Meta-learning
  instance identifiers for video object detection,'' \emph{IEEE Transactions on
  Image Processing}, vol.~30, pp. 6879--6891, 2021.

\bibitem{gong2021temporal}
T.~Gong, K.~Chen, X.~Wang, Q.~Chu, F.~Zhu, D.~Lin, N.~Yu, and H.~Feng,
  ``Temporal roi align for video object recognition,'' in \emph{Proceedings of
  the AAAI Conference on Artificial Intelligence}, vol.~35, no.~2, 2021, pp.
  1442--1450.

\bibitem{Cui_2021_ICCV}
Y.~Cui, L.~Yan, Z.~Cao, and D.~Liu, ``Tf-blender: Temporal feature blender for
  video object detection,'' in \emph{Proceedings of the IEEE/CVF International
  Conference on Computer Vision (ICCV)}, October 2021, pp. 8138--8147.

\bibitem{han2021cfanet}
L.~Han, P.~Wang, Z.~Yin, F.~Wang, and H.~Li, ``Class-aware feature aggregation
  network for video object detection,'' \emph{IEEE Transactions on Circuits and
  Systems for Video Technology}, pp. 1--1, 2021.

\bibitem{zhou2019objects}
X.~Zhou, D.~Wang, and P.~Kr{\"a}henb{\"u}hl, ``Objects as points,'' \emph{arXiv
  preprint arXiv:1904.07850}, 2019.

\bibitem{xu2020centernet}
Z.~Xu, E.~Hrustic, and D.~Vivet, ``Centernet heatmap propagation for real-time
  video object detection,'' in \emph{European Conference on Computer
  Vision}.\hskip 1em plus 0.5em minus 0.4em\relax Springer, 2020, pp. 220--234.

\end{thebibliography}

\clearpage
% \newpage

\end{document}